\documentclass[preprints,article,accept,moreauthors,unicode]{Definitions/mdpi}

\firstpage{1}
\pubvolume{1}
\issuenum{1}
\articlenumber{0}
\pubyear{2026}
\copyrightyear{2026}
\datereceived{}
\daterevised{}
\dateaccepted{}
\datepublished{}

\usepackage{catchfile}
\usepackage{centernot}
\usepackage[ruled,vlined,linesnumbered,algo2e]{algorithm2e}
\usepackage{float}
\usepackage{flafter}
\usepackage{placeins}
\DontPrintSemicolon
\SetKwComment{Comment}{\(\triangleright\) }{}
\usetikzlibrary{arrows.meta,positioning,calc,fit,backgrounds,decorations.pathreplacing}
\newcommand{\B}{\mathbb{B}}
\newcommand{\cB}{\mathcal{B}}

\newcommand{\Map}[1]{\ensuremath{M_{#1}}}
\newenvironment{theorem}{\begin{Theorem}}{\end{Theorem}}
\newenvironment{lemma}{\begin{Lemma}}{\end{Lemma}}
\newenvironment{proposition}{\begin{Proposition}}{\end{Proposition}}
\newenvironment{corollary}{\begin{Corollary}}{\end{Corollary}}
\newenvironment{remark}{\begin{Remark}}{\end{Remark}}

\Title{H-BRICK: Hierarchical Boundary Reachability Index with Compressed Kleene Closure for Directed Grid Graphs}

\Author{Tolga Özaslan $^{1,}$*\orcidA{}, Elif A. Özaslan $^{2}$\orcidB{}}

\AuthorNames{Tolga Özaslan and Elif A. Özaslan}

\address{%
$^{1}$ \quad Department of Computer Engineering, Ankara Yıldırım Beyazıt University, Ankara, Türkiye; tozaslan@aybu.edu.tr\\
$^{2}$ \quad Department of Mathematics, Kastamonu University, Kastamonu, Türkiye; eaozaslan@kastamonu.edu.tr}

\corres{Correspondence: tozaslan@aybu.edu.tr}

\abstract{
Repeated directed reachability queries arise in online path planning, task--agent allocation, and swarm control on static grid maps, where reachability is asymmetric and full transitive closure is storage-prohibitive.
We introduce H-BRICK, a Hierarchical Boundary Reachability Index with Compressed Kleene closure for exact cell-to-cell queries on directed grid graphs.
H-BRICK partitions the graph into tiles, computes a local Boolean Kleene closure within each tile, and projects it onto the tile's boundary ports.
The resulting directed port-to-port summaries are then composed recursively across shared interfaces up to a single root covering the graph.
We prove algebraically that the construction is exact: in the Boolean Kleene algebra of binary relations, every composed summary equals the boundary projection of the regional reflexive transitive closure, and every cell-to-cell query agrees with reachability in the original graph.
Experiments on directed grid graphs across diverse map topologies compare H-BRICK against directed search and component-based baselines, and compare its storage requirements against full transitive closure.
A batched reachability evaluation further demonstrates that sharing endpoint lifts across many-source/many-target queries amortizes hierarchical traversal and, at \(k=64\), yields a \(5.0\times\)--\(22.9\times\) throughput improvement over the scalar query loop.
The results show that hierarchical boundary compression is most advantageous in query-intensive spatial graphs where online search traverses large reachable regions and the hierarchy maintains compact boundary interfaces.
}



\begin{document}

\section{Introduction}

A reachability query asks whether a target vertex is reachable from a source vertex, and it arises in robotic navigation \cite{zhou2021raptor,gonzalez2020uav}, path planning \cite{paden2016survey,delling2017customizable}, map analysis \cite{wu2017reachable,sun2020frontier}, program analysis \cite{sui2016svf,shi2024pearl}, and graph database query processing \cite{hanauer2021oreach,zhang2025indexing}.
In grid-like environments such as occupancy maps, discretized workspaces, and structured road or terrain models, it must be answered repeatedly for many source--target pairs, including path-planning and task-assignment settings on rasterized warehouse maps \cite{Jiang2023GridMap}, such as candidate allocations in multi-agent pathfinding and swarm control \cite{thrun2005probabilistic,lavalle2006planning,stern2019mapf,ma2017lifelong,aakash2024makespan}.
This paper studies that setting in its directed form where each traversable cell is a vertex, and a directed edge is a feasible transition between neighboring cells.
Directionality may arise from one-way passages, flow fields, terrain-dependent motion constraints, traffic rules, or asymmetric motion primitives.
Reachability is therefore asymmetric: a path from \(u\) to \(v\) need not imply a path from \(v\) to \(u\).
Thus connected-component labeling settles the undirected case after a linear preprocess, but it does not suffice in this directed setting.

Classical solutions occupy a wide spectrum.
Running a graph search such as breadth-first search (BFS) for each query requires almost no preprocessing but repeatedly explores overlapping parts of a static graph, whereas materializing the all-pairs reachability relation \cite{warshall1962theorem,floyd1962algorithm,purdom1970transitive} yields constant-time lookup at a preprocessing and storage cost that becomes prohibitive once the closure grows dense.
Between these extremes, strongly connected component (SCC) contraction \cite{tarjan1972depth} resolves queries on the condensation, a directed acyclic graph whose vertices are the mutually reachable classes.
That reduction follows directed connectivity alone: an SCC may occupy an irregular region whose boundary has no planning meaning, whereas a path-planning map is usually organized into well-defined spatial tiles, often rectangular, whose sides are the natural interfaces between rooms, corridors, or map blocks.
The index developed here is geometry-driven instead, and exposes only boundary reachability to higher hierarchy levels.

We formalize reachability through Boolean matrices and Kleene algebra \cite{kozen1997kleene,armstrong2013kleene,guttmann2020relational,rote1990path,mohri2002semiring,hofner2012dijkstra,sergey2012calculating}, which H-BRICK applies locally by computing regional closures, projecting them onto compact boundary interfaces, and storing exact relations that cross-tile queries reuse.

Reachability indexing balances preprocessing time, index size, and query latency through precomputed intermediate relations \cite{yu2010graph,zhang2025indexing}.
Representative indexes include interval labeling \cite{agrawal1989efficient,cheng2006fast}, 2-hop labeling \cite{cohen2003reachability}, GRAIL \cite{yildirim2010grail}, Path-Tree \cite{jin2011path}, and O'Reach \cite{hanauer2021oreach}.
They target arbitrary large directed graphs, and they overlap with this paper on query type and workload which can be summarized as a single preprocess, then repeated Boolean path-existence queries between arbitrary vertices.
Their auxiliary structure is a vertex labeling or hop cover of a general digraph whereas this work uses the spatial boundary that a path-planning grid map already supplies in the form of tile sides, doorways, and corridor mouths.
Label-constrained and regular-path indexes \cite{zhang2023reachability,arroyuelo2022time,na2022regular,liang2024efficient,arroyuelo2024optimizing} lie outside the Boolean fragment used here.

Closest in spirit is border-based indexing, which cuts a graph into components and indexes connectivity through their border nodes \cite{schenkel2004hopi,cheng2006fast,cheng2008fast,jin2012scarab,gurajada2016distributed}.
Those constructions cover cross-partition connections via 2-hop labeling, reachability backbones, or explicit border graphs.
Related boundary abstractions appear in directed grid and layered-planar reachability algorithms \cite{jain2023solving,chakraborty2017directed,ashida2018reachability}, which replace subgrid paths with boundary edges to decide single queries within sublinear workspace.
Similarly, planar dense distance graphs \cite{fakcharoenphol2006planar} maintain metric shortest-path matrices over recursive $r$-divisions in the tropical semiring.
Unlike space-bounded decision algorithms that recompute auxiliary edges on the fly, H-BRICK targets the persistent repeated-query indexing regime.
Unlike metric or cost-bounded partition indexes \cite{tesfaye2022speeding,botea2004near}, H-BRICK formalizes boundary reachability over the Boolean Kleene semiring $(\mathbb{B}, \oplus, \otimes, 0, 1)$.
It presents a complete recursive hierarchy of exact boundary relations, precomputes dedicated rectangular boundary attachments ($R_T^{VB}, R_T^{BV}$, Eq.~\eqref{eq:base-attachment}) for arbitrary interior endpoints, and answers cell-to-cell queries via an ancestor scan ($z_U(s,t)$, Eq.~\eqref{eq:ancestor-test}) that detects paths leaving and re-entering smaller regions without fallback graph traversal in the canonical full-depth configuration.

Four-neighbor directed grids, obstacles included, are planar digraphs, so they lie in the class already solved by compact planar reachability oracles \cite{thorup2004compact,holm2015planar}.
Those oracles achieve near-linear preprocessing and space with constant-time queries via deep topological separator machinery; this work does not seek to improve their asymptotic worst-case bounds, but builds a cache-friendly, algebraically compositional index from local Kleene closures and grid boundaries, measuring the resulting empirical space--time tradeoff against tile granularity.
Hierarchical pathfinding (e.g., HPA* \cite{botea2004near,jansen2007hpa}) and contraction hierarchies \cite{geisberger2008contraction,delling2017customizable} similarly summarize spatial clusters by entrance distances to accelerate cost-driven search; in contrast, H-BRICK materializes exact Boolean boundary reachability for repeated path-existence testing where optimistic approximations are unacceptable.

Sparse Boolean linear algebra provides the computational substrate: GraphBLAS expresses graph algorithms over semirings \cite{kepner2016graphblas}, and high-performance bit-parallel kernels accelerate matrix multiplication and closure \cite{zhou2023bitwise,patel2020accelerating}.
These kernels supply primitive execution engines rather than index structures; our contribution lies in determining which boundary relations are materialized and how they compose.

We introduce H-BRICK, a Hierarchical Boundary Reachability Index with Compressed Kleene closure, which establishes a persistent rooted hierarchy for exact cell-to-cell reachability.
Each tile is closed locally and projected onto its boundary ports, and child summaries are composed recursively across shared interfaces up to a single root covering the graph, deciding arbitrary queries without materializing the complete \(O(N^2)\) closure.

The contributions are as follows.
(i) We formulate boundary-projected Boolean Kleene closure as a compositional framework and construct H-BRICK, a persistent rooted spatial hierarchy for exact cell-to-cell reachability on directed grid graphs.
(ii) We prove algebraically that every composed boundary summary equals the boundary projection of the regional reflexive transitive closure, so every cell-to-cell query agrees with reachability in the original graph.
(iii) We evaluate H-BRICK against directed search, SCC condensation, and labeling baselines across diverse topologies and directional flow regimes. Across 48 matched instances, its geometric-mean speedup over BFS is \(190\times\). On the largest map, bit-packed dense transitive closure requires about \(240\times\) the H-BRICK index storage at the selected \((b=24, g=4)\) operating point, while at \((b=32, g=4)\) the observed per-map max-to-min mean-query-latency ratios across orientation regimes range from \(1.38\times\) to \(6.24\times\).
(iv) We develop and evaluate a batched many-source/many-target reachability algorithm that amortizes hierarchical traversals, achieving, at \(k=64\), a \(5.0\times\)--\(22.9\times\) throughput improvement over scalar query loops for multi-agent workloads.

\section{Preliminaries, Problem Formulation, and Method Overview}
\label{sec:preliminaries}

\subsection{Directed Grid-Like Graphs and Reachability}

Let \(G=(V,E)\) be a static directed graph obtained from a grid map, such as the occupancy grid of an office or a warehouse or a discretized terrain, and write \(N=|V|\) and \(M=|E|\).
Each vertex is a traversable cell, and each directed edge \((u,v)\) is a feasible transition from \(u\) to a physically adjacent cell \(v\).
The edge relation need not be symmetric, so \((u,v)\in E\) does not imply \((v,u)\in E\).
A reachability query asks whether a directed path exists from a source \(s\in V\) to a target \(t\in V\), returning a Boolean value \(\operatorname{Reach}_G(s,t)\) indicating path existence without requiring path reconstruction or length optimization.
Zero-length paths are admitted, so \(\operatorname{Reach}_G(v,v)=1\) for every vertex \(v\in V\).

\subsection{Boolean Closure}

Reachability is expressed over the Boolean semiring \(\cB=(\B,\oplus,\otimes,0,1)\), where \(\B=\{0,1\}\) is the Boolean domain, addition \(\oplus\) is logical disjunction (with additive zero \(0\)), multiplication \(\otimes\) is logical conjunction (with multiplicative identity \(1\)), and multiplication distributes over addition.
Boolean matrix addition is entrywise, while Boolean matrix multiplication \(C=A\otimes B\) is defined by \(C_{ij}=\bigoplus_k(A_{ik}\otimes B_{kj})\).

Let \(A\in\B^{N\times N}\) be the Boolean adjacency matrix of \(G\), where \(A_{ij}=1\) if and only if \((i,j)\in E\).
Boolean powers encode directed walks of fixed length: \((A^{\otimes k})_{ij}=1\) if and only if there exists a directed walk of exactly \(k\) edges from \(i\) to \(j\).
The Boolean Kleene closure of \(A\) is
\begin{align}
    \label{eq:kleene-closure}
    A^\star = I \oplus A \oplus A^{\otimes 2} \oplus A^{\otimes 3} \oplus \cdots
    = I \oplus A \oplus A^{\otimes 2} \oplus \cdots \oplus A^{\otimes(N-1)},
\end{align}
where \(A^{\otimes k}\) denotes the product of \(k\) copies of \(A\) under Boolean matrix multiplication, and \(I_N\in\B^{N\times N}\) is the Boolean identity matrix with \(I_{ii}=1\) and \(I_{ij}=0\) for \(i\neq j\).
For notational simplicity, where no ambiguity arises, we subsequently write \(A^k\) interchangeably for the Boolean power \(A^{\otimes k}\).
The sum truncates at \(A^{N-1}\) because every simple directed path in \(G\) contains at most \(N-1\) edges.
Boolean matrices over \(\cB\), equipped with entrywise \(\oplus\), Boolean matrix multiplication \(\otimes\), and Kleene closure, form the Kleene algebra \((\B^{N\times N},\oplus,\otimes,(\cdot)^\star,0_{N\times N},I_N)\) of binary relations on \(V\), where \(\oplus\) denotes relational union, \(\otimes\) denotes relational composition, and \((\cdot)^\star\) denotes reflexive transitive closure \cite{Krivulin2021Algebraic,Deng2026TransitiveClosure}.
Consequently, \((A^\star)_{ij}=1\) if and only if there exists a directed path of length \(k\ge 0\) from \(i\) to \(j\), making \(A^\star\) the reflexive transitive closure of \(G\) in matrix form.

It is also significant that because \(\oplus\) is idempotent,
\begin{align*}
    (I\oplus A)^2 & = I\oplus A\oplus A\oplus A^2 = I\oplus A\oplus A^2,                                               \\
    (I\oplus A)^4 & = (I\oplus A\oplus A^2)^2 = I\oplus A\oplus A^2\oplus A^3\oplus A^4,
\end{align*}
so \(A^\star=(I\oplus A)^{N-1}=(I\oplus A)^{2^{\lceil\log_2(N-1)\rceil}}\) for \(N\ge 2\), which repeated squaring of \(I\oplus A\) reaches in \(\lceil\log_2(N-1)\rceil\) Boolean matrix multiplications.

\subsection{Spatial Tiles and Boundary Ports}

In the two-dimensional grid, a tile \(T\) is a rectangular block of cells.
Let \(V_T\subseteq V\) be the vertices of \(T\), and write \(n_T=|V_T|\).
The induced subgraph is \(G_T=(V_T,E_T)\) with \(E_T=E\cap(V_T\times V_T)\), whose Boolean adjacency matrix is \(A_T\in\B^{n_T\times n_T}\).

Formally, the boundary port set \(B_T\) consists of all passable cells of \(T\) incident to an edge crossing between \(T\) and the remainder of the graph:
\begin{align}
    \label{eq:port-set}
    B_T = \left\{ u\in V_T : \exists v\in V\setminus V_T \text{ such that } (u,v)\in E \text{ or } (v,u)\in E \right\},
\end{align}
ordered canonically with \(b_T=|B_T|\).
Each such boundary cell acts as an interface port (or simply port) mediating inter-tile transitions in the global reachability graph.
The exposed summary \(S_T\in\B^{b_T\times b_T}\) is the Boolean relation on \(B_T\) with \((S_T)_{pq}=1\) exactly when \(q\) is reachable from \(p\) by a directed path contained in \(T\).
On the four-neighbor rectangular grids considered here, any vertex incident to an external edge necessarily lies on the perimeter of \(T\).
Defining \(B_T\) via~\eqref{eq:port-set} automatically admits both incoming-only and outgoing-only ports while pruning inactive perimeter cells, minimizing \(b_T\) (with \(B_r=\varnothing\) at the root region \(r\) covering \(V\)).

\subsection{Problem Formulation and Method Overview}
\label{subsec:interfaces}

The query workload on \(G\) is an arbitrary sequence of pairs \((s_i,t_i) \in V \times V\), each evaluated for \(\operatorname{Reach}_G(s_i,t_i)\).
A rooted spatial decomposition \(\mathcal{T}\) is valid if its leaves partition \(V\), every internal node is the union of its children, boundary-crossing edges are incident to declared ports, and the root covers \(V\).

\textbf{Problem 1 (Repeated Directed Reachability Indexing).}
Given a static directed grid-like graph \(G=(V,E)\) and a valid rooted spatial decomposition \(\mathcal{T}\), construct an index \(\mathcal{I}(G,\mathcal{T})\) satisfying
\begin{align}
    \label{eq:exactness-requirement}
    \operatorname{Query}_{\mathcal{I}}(s,t) = \operatorname{Reach}_G(s,t) \quad \text{for all } (s,t)\in V\times V.
\end{align}

In this paper, \(\mathcal{I}(G,\mathcal{T})\) is realized by H-BRICK.
Each base tile is closed locally, storing attachments that record reachability from tile vertices to boundary ports and from boundary ports to tile vertices, and exposing only the induced reachability relation over those ports.
At each higher level, child summaries and directed seam transitions form a reduced interface graph.
Each parent retains the closure over the interface ports of its children for query processing and exposes the restriction of that closure to its own external ports for composition at the next level.
Composition continues recursively to a unique root covering \(G\), where no external boundary remains.
The formulation accommodates arbitrary tile dimensions and recursive grouping factors.

\begin{figure}[!t]
\centering
\resizebox{0.96\linewidth}{!}{%
\input{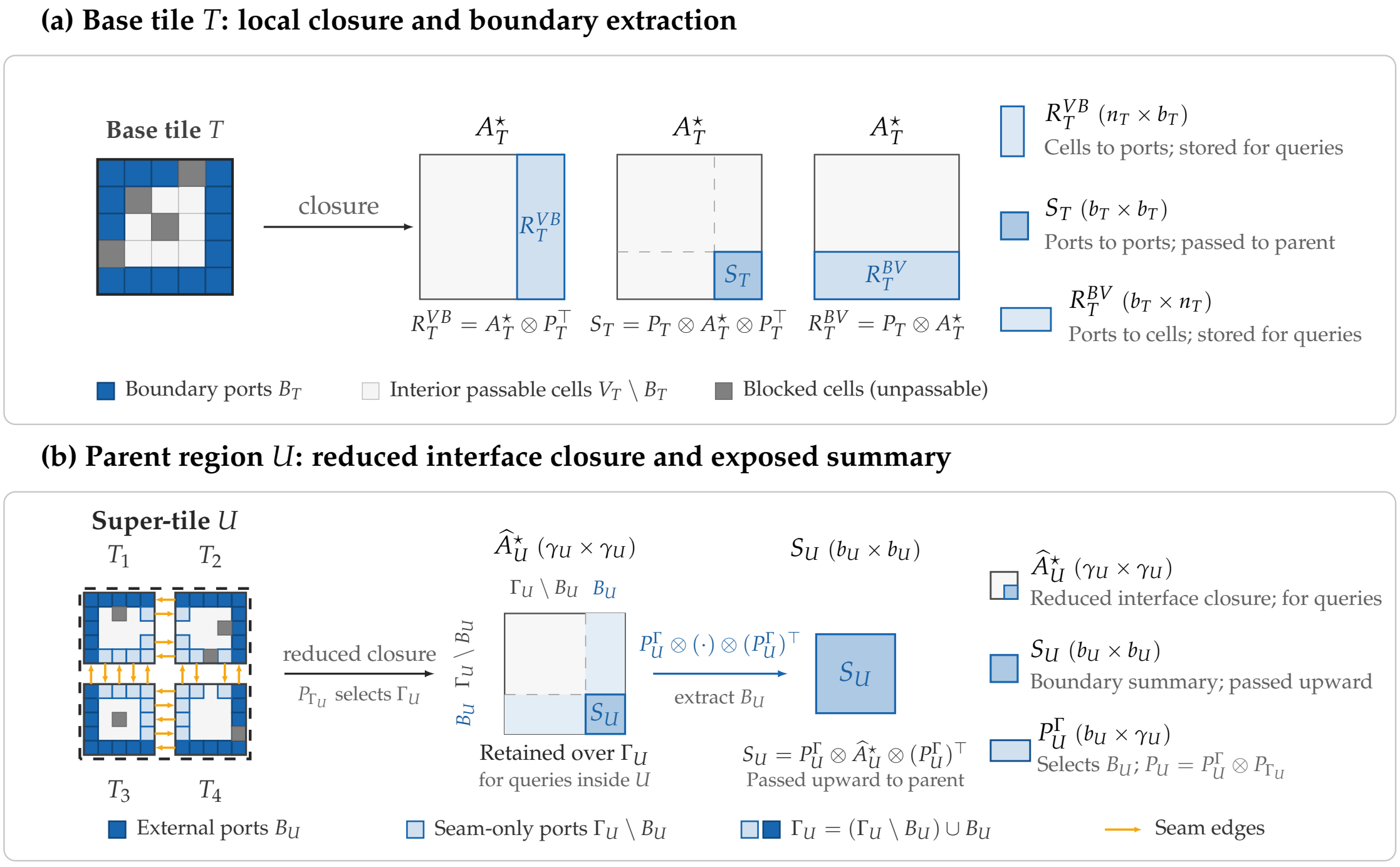}%
}
\caption{Structural matrix roles at a base tile and at a parent.
(\textbf{a}) Base tile \(T\): boundary selector \(P_T\) extracts query attachments \(R_T^{VB}, R_T^{BV}\) and summary \(S_T\) from local closure \(A_T^\star\).
(\textbf{b}) Parent region \(U\): oblivious to child interiors (\(T_1,\ldots,T_4\)), \(U\) combines child summaries and seam edges into interface closure \(\widehat{A}_U^\star\), extracting summary \(S_U\) via \(P_U^\Gamma\) to pass upward. The selector \(P_{\Gamma_U}\) selects \(\Gamma_U\) from \(V_U\) strictly as an expository proof construct (Theorem~\ref{thm:composition}) and is never materialized or executed.}
\label{fig:matrix-roles}%
\makeatletter
\def\@currentlabel{\thefigure a}\label{fig:matrix-roles-a}%
\def\@currentlabel{\thefigure b}\label{fig:matrix-roles-b}%
\makeatother
\end{figure}

Queries read stored relations without recomputing closures.
When source \(s\) and target \(t\) occupy the same base tile, the local closure is inspected directly.
Otherwise, source-to-port and port-to-target attachments are combined with stored interface closures across common ancestors, which is necessary because directed paths may exit a smaller region and re-enter it through external seams.
Table~\ref{tab:notation} summarizes the notation used in Sections~2--4, and Figure~\ref{fig:matrix-roles} diagrams the matrix pipeline from a base tile through parent composition.

\begin{table}[!t]
\begin{adjustwidth}{-\extralength}{0cm}
    \caption{Core notation. All matrices and vectors are Boolean; dimensions are rows \(\times\) columns. Adjacency and reachability matrices use the row-source, column-target convention.}
    \label{tab:notation}
    \centering
    \footnotesize
    \setlength{\tabcolsep}{4pt}
    \renewcommand{\arraystretch}{1.18}
    \begin{tabular*}{\fulllength}{@{\extracolsep{\fill}}l r l@{}}
        \hline
        \rule{0pt}{2.4ex}Symbol                                         & Size                         & Brief description                                                                                          \\[0.5ex]
        \hline
        \(\mathbb{B}\)                                                  & ---                          & Boolean domain \(\{0,1\}\).                                                                                \\
        \(\oplus,\otimes\)                                              & ---                          & Scalar OR and AND; matrix entrywise OR and Boolean multiplication.                                         \\
        \(\mathcal{B}\)                                                 & ---                          & Boolean semiring \((\mathbb{B},\oplus,\otimes,0,1)\).                                                      \\
        \((\cdot)^\star\)                                               & ---                          & Reflexive transitive closure of a Boolean square matrix.                                                   \\
        \(G=(V,E),\,N,\,M\)                                             & ---                          & Directed cell graph; \(N=\lvert V\rvert\), \(M=\lvert E\rvert\).                                           \\
        \rule{0pt}{2.5ex}\(A,\,A^\star\)                                & \(N\times N\)                & Adjacency matrix of \(G\) and its closure (Eq.~\eqref{eq:kleene-closure}).                                 \\
        \hline
        \(T,\,V_T,\,n_T\)                                               & ---                          & Tile, its traversable vertices, and \(n_T=\lvert V_T\rvert\).                                              \\
        \(\mathcal{T},\,r\)                                             & ---                          & Rooted hierarchy of tiles; root \(r\) covers \(V\).                                                        \\
        \(U,\,\operatorname{ch}(U),\,m\)                                & ---                          & Parent tile with children \(\operatorname{ch}(U)=\{T_1,\ldots,T_m\}\).                                     \\
        \(B_T,\,b_T\)                                                   & ---                          & Vertices of \(T\) incident to crossing edges; \(b_T=\lvert B_T\rvert\) (Eq.~\eqref{eq:port-set}).          \\
        \rule{0pt}{2.5ex}\(A_T,\,A_T^\star\)                            & \(n_T\times n_T\)            & Adjacency of subgraph induced by \(V_T\), and its closure (Eq.~\eqref{eq:kleene-closure}).                 \\
        \(P_T\)                                                         & \(b_T\times n_T\)            & Selects boundary ports \(B_T\) from vertex ordering \(V_T\) (Section~\ref{sec:boundary-kleene}).           \\
        \(S_T\)                                                         & \(b_T\times b_T\)            & Exposed summary: reachability between ports inside \(T\) (Eq.~\eqref{eq:base-summary}).                    \\
        \rule{0pt}{2.5ex}\(R_T^{VB}\)                                   & \(n_T\times b_T\)            & Vertex-to-port reachability inside base tile \(T\) (Eq.~\eqref{eq:base-attachment}).                       \\
        \rule{0pt}{2.5ex}\(R_T^{BV}\)                                   & \(b_T\times n_T\)            & Port-to-vertex reachability inside base tile \(T\) (Eq.~\eqref{eq:base-attachment}).                       \\
        \hline
        \(\Gamma_U,\,\gamma_U\)                                         & ---                          & Ordered union of child port sets \(B_{T_i}\); \(\gamma_U=\lvert\Gamma_U\rvert\).                           \\
        \(P_{\Gamma_U}\)                                                & \(\gamma_U\times n_U\)       & Selects child-interface ports \(\Gamma_U\) from \(V_U\) (Eq.~\eqref{eq:composition-exactness}).            \\
        \rule{0pt}{2.5ex}\(J_i\)\rule[-0.7ex]{0pt}{0pt}                 & \(\gamma_U\times b_{T_i}\)   & Embeds child ports \(B_{T_i}\) into ordering \(\Gamma_U\) (Eq.~\eqref{eq:composition}).                    \\
        \rule{0pt}{2.6ex}\(A_U^{\mathrm{seam}}\)                        & \(\gamma_U\times\gamma_U\)   & Adjacency of directed edges crossing between children of \(U\) (Eq.~\eqref{eq:composition}).               \\
        \rule{0pt}{2.6ex}\(\widehat{A}_U\)\rule[-0.8ex]{0pt}{0pt}       & \(\gamma_U\times\gamma_U\)   & Reduced interface adjacency combining summaries and seams (Eq.~\eqref{eq:composition}).                    \\
        \rule{0pt}{2.8ex}\(\widehat{A}_U^\star\)\rule[-0.9ex]{0pt}{0pt} & \(\gamma_U\times\gamma_U\)   & Reachability inside \(U\) among child-interface ports \(\Gamma_U\) (Eq.~\eqref{eq:composition-exactness}). \\
        \rule{0pt}{2.5ex}\(P_U^\Gamma\)                                 & \(b_U\times\gamma_U\)        & Selects external parent ports \(B_U\) from \(\Gamma_U\) (Eq.~\eqref{eq:parent-summary}).                   \\
        \hline
        \(\mathcal{I}(G,\mathcal{T})\)                                  & ---                          & H-BRICK reachability index for \(G\) and its tile hierarchy.                                               \\
        \(s,t,\,T_s,T_t\)                                               & ---                          & Query source and target, and their respective base tiles.                                                  \\
        \(W\)                                                           & ---                          & Ancestor in one-step frontier propagation; common ancestor \(U\) in \(z_U(s,t)\). \\
        \(C_s(W),\,C_t(W)\)                                             & ---                          & Child of \(W\) containing \(s\), respectively \(t\).                                                       \\
        \rule{0pt}{2.5ex}\(x_W(s),\,y_W(t)\)\rule[-0.7ex]{0pt}{0pt}     & \(1\times b_W,\ b_W\times1\) & Source-to-port and port-to-target reachability inside \(W\) (Eq.~\eqref{eq:query-propagation}).            \\
        \rule{0pt}{2.4ex}\(z_U(s,t)\)\rule[-0.6ex]{0pt}{0pt}            & Scalar                       & Boolean connection test through interface of common ancestor \(U\) (Eq.~\eqref{eq:ancestor-test}).         \\
        \hline
    \end{tabular*}
\end{adjustwidth}
\end{table}

\section{Boundary-Compressed Kleene Closure}
\label{sec:boundary-kleene}

This section defines the boundary-projected closure and attachment relations of a base tile; Section~\ref{sec:hierarchy} extends the construction recursively.
Fix a local vertex ordering of \(V_T\) and a deterministic ordering of the distinct ports in \(B_T\).
Define the boundary selector \(P_T\in\mathbb{B}^{b_T\times n_T}\) by \((P_T)_{ij}=1\) exactly when the \(i\)-th boundary port is the \(j\)-th local vertex.
This admits arbitrary local vertex orderings, with inter-tile mappings determined by vertex identities.

For a base tile \(T\), its local closure \(A_T^\star\) is computed according to~\eqref{eq:kleene-closure} via the bit-parallel closure engine (Section~\ref{sec:analysis}), and projected onto its boundary ports \(B_T\):
\begin{align}
    \label{eq:base-summary}
    S_T
     & =
    P_T
    \otimes
    A_T^\star
    \otimes
    P_T^{\top}
    \in\mathbb{B}^{b_T\times b_T}.
\end{align}

\begin{lemma}[Local boundary summary correctness]
    \label{lem:local-summary}
    Let \(T\) be a base tile with local adjacency matrix \(A_T\), boundary selector \(P_T\), and summary \(S_T\) given by \eqref{eq:base-summary}.
    For boundary ports \(p,q\in B_T\), \((S_T)_{pq}=1\) if and only if \(q\) is reachable from \(p\) by a directed path contained entirely in \(T\).
\end{lemma}

\begin{proof}
    Because \(A_T^\star\) is the reachability relation on \(V_T\), left and right multiplication by \(P_T\) restricts its coordinates to \(B_T\), yielding exactly the boundary reachability relation.
\end{proof}

Because query endpoints may lie strictly in the interior of a base tile, each base tile also stores the attachment relations
\begin{align}
    \label{eq:base-attachment}
    R_T^{VB}
     & =
    A_T^\star
    \otimes
    P_T^{\top}
    \in\mathbb{B}^{n_T\times b_T},
     &
    R_T^{BV}
     & =
    P_T
    \otimes
    A_T^\star
    \in\mathbb{B}^{b_T\times n_T}.
\end{align}

\begin{lemma}[Attachment correctness]
    \label{lem:attachment}
    For a base tile \(T\) with the attachment relations of \eqref{eq:base-attachment}, \((R_T^{VB})_{sp}=1\) if and only if boundary port \(p\) is reachable from local vertex \(s\) inside \(T\), and \((R_T^{BV})_{qt}=1\) if and only if local vertex \(t\) is reachable from boundary port \(q\) inside \(T\).
\end{lemma}

\begin{proof}
    Under the row-source/column-target convention, right multiplication by \(P_T^{\top}\) selects boundary target columns and left multiplication by \(P_T\) selects boundary source rows, giving vertex-to-port and port-to-vertex reachability, respectively.
    In directed graphs, these relations are not transposes of one another.
\end{proof}

The implementation retains the full local closure \(A_T^\star\) for direct same-tile queries, while materializing the boundary summary \(S_T\) for hierarchical composition and the attachment relations \(R_T^{VB}\) and \(R_T^{BV}\) for cross-tile query evaluation (Figure~\ref{fig:matrix-roles-a}).

\section{H-BRICK Hierarchical Composition and Exact Querying}
\label{sec:hierarchy}

This section composes base-tile summaries into super-tile summaries \(S_U\), builds a rooted spatial hierarchy covering \(V\), and answers arbitrary cell-to-cell queries.

\subsection{Parent Interface Space}

\begin{figure}[!t]
    \centering
    \resizebox{\linewidth}{!}{%
        \input{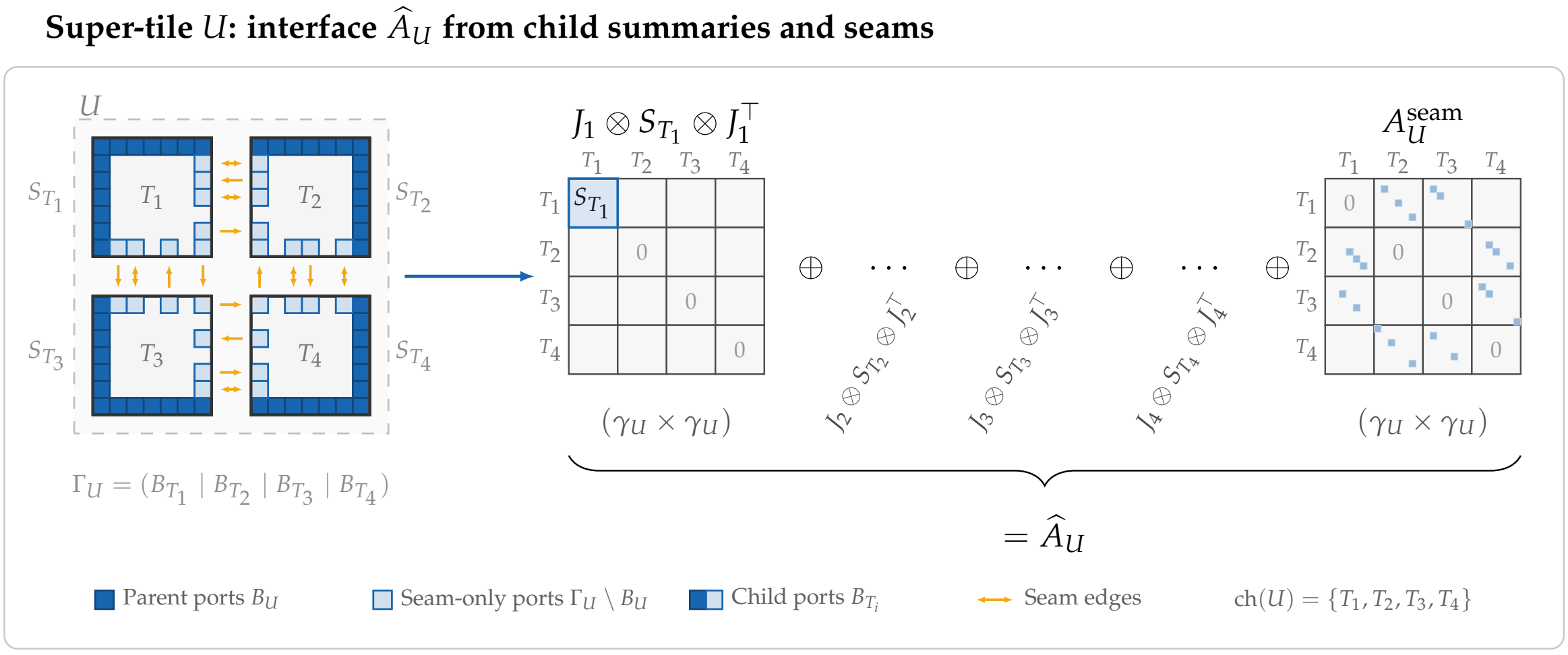}%
    }
    \caption{Parent interface of a super-tile \(U\) formed by children \(\operatorname{ch}(U)=\{T_1,T_2,T_3,T_4\}\).
    Child summaries \(S_{T_i}\) are embedded into interface \(\Gamma_U\) via \(J_i\), and combined with directed seam transitions \(A_U^{\mathrm{seam}}\) (gold arrows) according to~\eqref{eq:composition}.
    Solid perimeter cells denote external ports \(B_U\subseteq\Gamma_U\), while light cells denote internal seam-only ports in \(\Gamma_U\setminus B_U\).}
    \label{fig:parent-interface}
\end{figure}

Let \(U\) be a super-tile formed from the child tiles \(\operatorname{ch}(U)=\{T_1,\ldots,T_m\}\), each of which exposes its boundary ports \(B_{T_i}\) and its already-computed summary \(S_{T_i}\).
Let \(\Gamma_U\) denote the ordered set of all boundary ports of the child tiles (which may each be a base tile or another super-tile) participating in the composition of \(U\), and write \(\gamma_U=|\Gamma_U|\).
Since the children are vertex-disjoint, every child port enters \(\Gamma_U\) under its own identity, and adjacency between two children is carried by seam edges.
For each child, a Boolean embedding matrix \(J_i \in \mathbb{B}^{\gamma_U \times b_{T_i}}\) maps the local port ordering of \(T_i\) into \(\Gamma_U\), and the embedded contribution of that child to the reduced interface graph is \(J_i \otimes S_{T_i} \otimes J_i^{\top}\in\mathbb{B}^{\gamma_U\times\gamma_U}\).
Here, the interface graph is termed ``reduced'' because it completely abstracts away all interior cells of the child regions, retaining only the child boundary ports \(\Gamma_U\) as its active vertices.
The seam adjacency \(A_U^{\mathrm{seam}}\in\mathbb{B}^{\gamma_U\times\gamma_U}\) records one-step directed edges of \(G\) crossing between ports of distinct children, while remaining \(0\) on every same-child pair (Figure~\ref{fig:parent-interface}).

\subsection{Composition}

The reduced interface adjacency matrix, \(\widehat{A}_U\), of \(U\) collects the embedded child summaries together with the seam edges of the interface relation, as drawn for four children in Figure~\ref{fig:parent-interface}:
\begin{align}
    \label{eq:composition}
    \widehat{A}_U
     & =
    \left(
    \bigoplus_{i=1}^{m}
    J_i
    \otimes
    S_{T_i}
    \otimes
    J_i^{\top}
    \right)
    \oplus
    A_U^{\mathrm{seam}}
    \in\mathbb{B}^{\gamma_U\times\gamma_U}.
\end{align}
Because the graph is directed, \(A_U^{\mathrm{seam}}\) is not symmetric unless both directions already exist in \(G\).
Same-child-port reachability enters \(\widehat{A}_U\) through \(J_i\otimes S_{T_i}\otimes J_i^{\top}\), whereas one-step transitions between distinct children enter through \(A_U^{\mathrm{seam}}\).
An edge of \(\widehat{A}_U\) is therefore either a child-internal connection or a one-step inter-child seam edge.
Closing each child does not close that union since a path may, one or multiple times, leave one child, cross a seam, and enter another, which is why the parent computes \(\widehat{A}_U^\star\).

\subsection{Parent Closure and Boundary Projection}

Closing the reduced interface graph gives \(\widehat{A}_U^\star\in\mathbb{B}^{\gamma_U\times\gamma_U}\), which records reachability among all of \(\Gamma_U\), including the external ports \(B_U\subseteq\Gamma_U\), using only child summaries and seam edges.
Write \(b_U=|B_U|\).
The set \(\Gamma_U\setminus B_U\) consists of seam-only child ports, shown as light cells in Figure~\ref{fig:parent-interface}, while a seam endpoint on the outer perimeter remains in \(B_U\) only if it is incident to an edge entering or leaving \(U\).
After that closure, \(P_U^\Gamma \in \mathbb{B}^{b_U \times \gamma_U}\) selects \(B_U\) from \(\Gamma_U\), and the exposed summary is the same projection used at the base level,
\begin{align}
    \label{eq:parent-summary}
    S_U
     & =
    P_U^\Gamma
    \otimes
    \widehat{A}_U^\star
    \otimes
    (P_U^\Gamma)^{\top}
    \in\mathbb{B}^{b_U\times b_U},
\end{align}
after which \(U\) is treated as an ordinary tile one level higher.
The construction is designed to preserve the level invariant
\begin{align}
    \label{eq:level-invariant}
    S_U
     & =
    P_U
    \otimes
    A_U^\star
    \otimes
    P_U^{\top}
    \in\mathbb{B}^{b_U\times b_U},
\end{align}
where \(A_U \in \mathbb{B}^{n_U \times n_U}\) is the full Boolean adjacency matrix of the subgraph covered by \(U\) (with \(n_U = |V_U|\)), \(A_U^\star \in \mathbb{B}^{n_U \times n_U}\) is its reflexive transitive closure, and \(P_U \in \mathbb{B}^{b_U \times n_U}\) is the Boolean selector extracting the external boundary ports \(B_U\) from the full vertex set of that subgraph.
The matrix \(A_U^\star\) is indexed by every vertex of \(U\) and is not computed; H-BRICK obtains the same boundary relation through \(\widehat{A}_U^\star\) (Figure~\ref{fig:matrix-roles-b}).

\begin{theorem}[Interface-closure exactness]
    \label{thm:composition}
    Let \(U\) be a super-tile formed from children \(\operatorname{ch}(U)=\{T_1,\ldots,T_m\}\) whose summaries \(S_{T_i}\) are exact, let \(A_U \in \mathbb{B}^{n_U \times n_U}\) be the full adjacency matrix of the subgraph induced by \(U\), let \(P_{\Gamma_U} \in \mathbb{B}^{\gamma_U \times n_U}\) select the interface ports \(\Gamma_U\) from the full vertex ordering of that subgraph, and let \(\widehat{A}_U \in \mathbb{B}^{\gamma_U \times \gamma_U}\) be the reduced interface adjacency matrix \eqref{eq:composition}.
    Then the interface closure computed from the reduced interface graph equals the interface projection of the full closure:
    \begin{align}
        \label{eq:composition-exactness}
        \widehat{A}_U^\star
         & =
        P_{\Gamma_U}
        \otimes
        A_U^\star
        \otimes
        P_{\Gamma_U}^{\top}
        \in\mathbb{B}^{\gamma_U\times\gamma_U}.
    \end{align}
\end{theorem}

\begin{proof}
    Both sides of \eqref{eq:composition-exactness}, \(\widehat{A}_U^\star\) and \(P_{\Gamma_U}\otimes A_U^\star\otimes P_{\Gamma_U}^{\top}\), are indexed by \(\Gamma_U\times\Gamma_U\).
    For any interface ports \(p,q\in\Gamma_U\) we show two directions: a directed path from \(p\) to \(q\) in the subgraph of \(U\) gives \((\widehat{A}_U^\star)_{pq}=1\), and a directed path from \(p\) to \(q\) in the reduced interface graph gives \((P_{\Gamma_U}\otimes A_U^\star\otimes P_{\Gamma_U}^{\top})_{pq}=1\).

    Suppose first that \(q\) is reachable from \(p\) by a directed path in the subgraph covered by \(U\).
    A seam crossing is a directed edge of \(G\) whose ends lie in two different children.
    Let's cut the path at every seam crossing.
    Then the resulting path alternates intra-child pieces with seams,
    \[
        (\text{inside }T_{i_1}),\
        \text{seam},\
        (\text{inside }T_{i_2}),\
        \text{seam},\
        \ldots,\
        (\text{inside }T_{i_k}).
    \]
    Each intra-child piece lies entirely in one child \(T_i\), with both endpoints in \(B_{T_i}\subseteq\Gamma_U\).
    Every nonfinal piece is followed by a seam edge leaving that child.
    Exactness of \(S_{T_i}\) records that piece of the path as a single entry of \(S_{T_i}\), not as a chain of cell hops, and \(J_i\) places the entry into \(\Gamma_U\).
    Each seam between two intra-child pieces is an entry of \(A_U^{\mathrm{seam}}\).
    The original path therefore gives a directed walk from \(p\) to \(q\) in the reduced interface graph, so \((\widehat{A}_U^\star)_{pq}=1\).

    Conversely, consider a directed \(p\)-\(q\) path in the reduced interface graph.
    Each of its edges is either an interface relation from \(A_U^{\mathrm{seam}}\), which corresponds to an actual directed transition of the full graph model of \(U\), or an embedded child-summary edge \(J_i \otimes S_{T_i} \otimes J_i^{\top}\), which by exactness of \(S_{T_i}\) corresponds to an actual directed path inside \(T_i\).
    Replacing every reduced edge by its corresponding inter-child seam edge or child-internal path yields a directed walk from \(p\) to \(q\) in the full subgraph induced by \(U\), hence a directed path, which is \((P_{\Gamma_U} \otimes A_U^\star \otimes P_{\Gamma_U}^{\top})_{pq}=1\).
    The two implications give \eqref{eq:composition-exactness}.
    Figure~\ref{fig:parent-interface} illustrates parts of the proof.
\end{proof}

\begin{corollary}[Exposed boundary summary]
    \label{cor:external-summary}
    Under the hypotheses of Theorem~\ref{thm:composition}, the exposed summary \eqref{eq:parent-summary} satisfies the level invariant \eqref{eq:level-invariant}, that is, \(S_U = P_U \otimes A_U^\star \otimes P_U^{\top}\), where \(P_U\) selects the external boundary ports of \(U\) from the full vertex ordering.
\end{corollary}

\begin{proof}
    The external ports satisfy \(B_U \subseteq \Gamma_U\), and selecting them from the full vertex ordering factors as \(P_U = P_U^\Gamma \otimes P_{\Gamma_U}\).
    Multiplying \eqref{eq:composition-exactness} by \(P_U^\Gamma\) on the left and by \((P_U^\Gamma)^{\top}\) on the right therefore gives \(S_U = P_U^\Gamma \otimes \widehat{A}_U^\star \otimes (P_U^\Gamma)^{\top} = P_U \otimes A_U^\star \otimes P_U^{\top}\).
\end{proof}

\subsection{Hierarchy Construction}

\begin{algorithm2e}[t]
    \caption{Super-Tile Summary Construction}
    \label{alg:composition}
    \KwIn{Parent tile \(U\), child summaries \(S_{T_1},\ldots,S_{T_m}\), child port sets \(B_{T_1},\ldots,B_{T_m}\), seams \(A_U^{\mathrm{seam}}\), external ports \(B_U\)}
    \KwOut{Interface closure \(\widehat{A}_U^\star\in\mathbb{B}^{\gamma_U\times\gamma_U}\); if \(U\neq r\), also the exposed summary \(S_U\in\mathbb{B}^{b_U\times b_U}\)}
    Construct the common ordered interface port set \(\Gamma_U\)\;
    \For{each child tile \(T_i\)}{
    Construct the embedding matrix \(J_i\in\mathbb{B}^{\gamma_U\times b_{T_i}}\)\;
    Embed the child summary as \(J_i \otimes S_{T_i} \otimes J_i^{\top}\)\;
    }
    Construct \(\widehat{A}_U \leftarrow \bigl(\bigoplus_{i=1}^{m} J_i \otimes S_{T_i} \otimes J_i^{\top}\bigr) \oplus A_U^{\mathrm{seam}}\)\;
    Compute \(\widehat{A}_U^\star\)\;
    \If{\(U=r\)}{\Return{\(\widehat{A}_U^\star\)}}
    Construct the external-boundary selection matrix \(P_U^\Gamma\in\mathbb{B}^{b_U\times\gamma_U}\)\;
    Compute \(S_U \leftarrow P_U^\Gamma \otimes \widehat{A}_U^\star \otimes (P_U^\Gamma)^{\top}\)\;
    \Return{\(\widehat{A}_U^\star,S_U\)}
\end{algorithm2e}

H-BRICK is a rooted spatial hierarchy covering \(V\).
Let \(\mathcal{T}\) be a tree whose leaves \(\mathcal{T}^{(0)}\) are the base tiles of \(G\).
Every internal node \(U\) (a non-leaf of \(\mathcal{T}\)) is the union of its children and stores the interface \(\Gamma_U\) together with \(\widehat{A}_U^\star\); the root \(r\) is the unique node whose region is the entire vertex set \(V\).
Under \eqref{eq:port-set}, \(B_r=\varnothing\) because \(r\) covers \(V\), so the index does not expose an external root summary \(S_r\); cell-to-cell queries use the reduced interface closure \(\widehat{A}_r^\star\).
Every non-root tile has exactly one parent.

Construction is bottom-up: equations~\eqref{eq:base-summary}--\eqref{eq:base-attachment} compute every leaf summary, then Algorithm~\ref{alg:composition} is applied to every internal node.
A nonroot parent stores the projection \(S_U\) of \eqref{eq:parent-summary}; the root retains and returns only \(\widehat{A}_r^\star\), obtained by closing \eqref{eq:composition} at \(U=r\).
If \(\mathcal{T}\) consists of a single base tile, it is the root and no composition is invoked.
The depth of \(\mathcal{T}\) is determined by the base-tile granularity and by the aggregation rule.


\begin{theorem}[Hierarchical exactness]
    \label{thm:hierarchical}
    Theorem~\ref{thm:composition} and Corollary~\ref{cor:external-summary} extend from one parent to every node of \(\mathcal{T}\):
    every nonroot tile \(T\) stores an exposed summary satisfying the level invariant \eqref{eq:level-invariant},
    and every internal node \(U\), including the root, stores an interface closure satisfying \eqref{eq:composition-exactness}.
\end{theorem}

\begin{proof}
    We prove the two identities by induction on the nodes of \(\mathcal{T}\), starting at the leaves and treating a parent only after all of its children are known.
    At a leaf, the exposed-summary identity is Lemma~\ref{lem:local-summary}, and \eqref{eq:composition-exactness} is vacuous, because a leaf has no child-port interface.
    Assume that every child of an internal node \(U\) has an exact exposed summary.
    Theorem~\ref{thm:composition} then gives \eqref{eq:composition-exactness} for \(U\).
    If \(U\) is not the root, Corollary~\ref{cor:external-summary} gives \eqref{eq:level-invariant} for \(U\).
    By induction every nonroot tile stores an exact exposed summary, and every internal node, including the root, stores an exact interface closure.
\end{proof}

Nothing in \eqref{eq:composition}--\eqref{eq:parent-summary} refers to tile shape: the operations depend only on the vertex set of a region, its exposed ports, and its seam relations.
Square and rectangular tiles are convenient for grid maps because they follow the grid geometry and admit simple side-based port orderings; the implementation evaluated below uses them throughout.

\subsection{Exact Cell-to-Cell Query}

The external query model is ordinary reachability between arbitrary vertices, so the source \(s\) and the target \(t\) need not be boundary ports.
Let \(T_s\) and \(T_t\) be the leaf tiles containing $s$ and $t$, respectively.
If \(T_s=T_t\), the query is first tested locally through \((A_{T_s}^\star)_{st}=1\).
A positive local test answers the query.
A negative local test is \emph{inconclusive}, because a valid path may leave the tile and re-enter it through a larger ancestor.
If the hierarchy is a single base tile, that tile is the root, there are no internal ancestors, and the local closure is the whole answer.

\begin{figure}[!t]
    \centering
    \resizebox{\linewidth}{!}{%
        \input{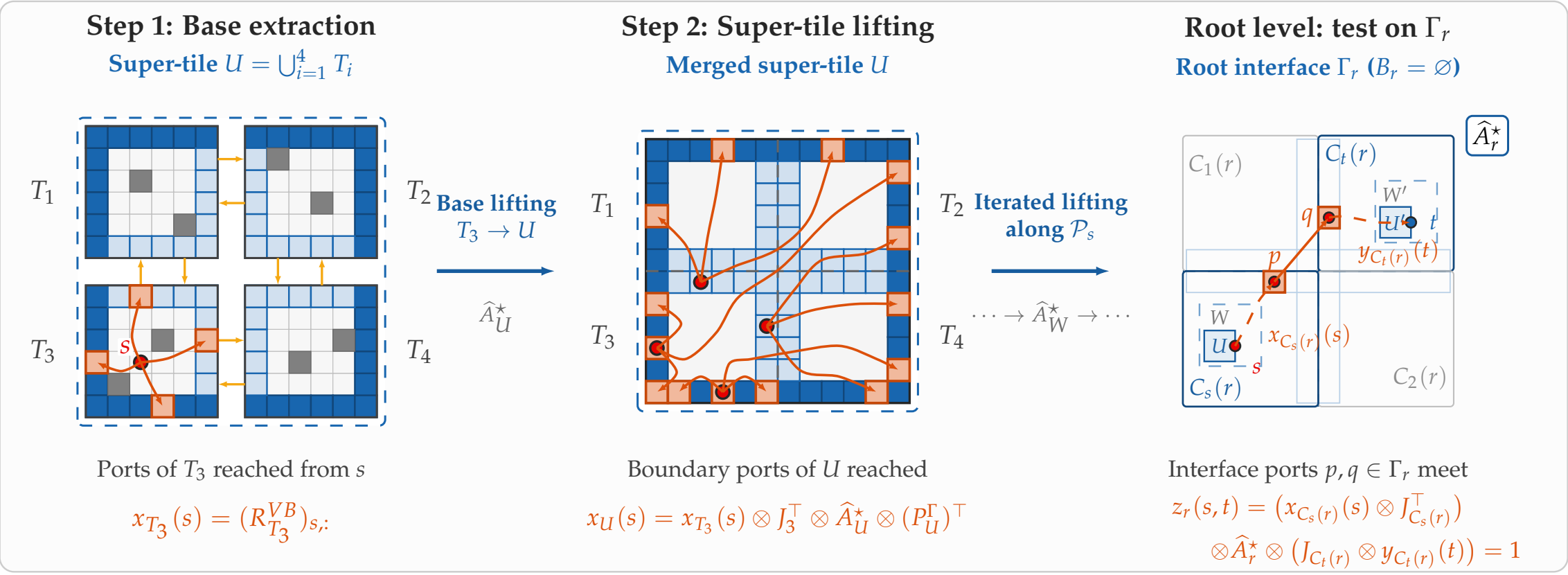}%
    }
    \caption{Hierarchical query vector lifting.
        In Step~1, source \(s\) reaches base-tile boundary ports \(x_{T_3}(s)\).
        In Step~2, active ports (red dots) are embedded into \(\Gamma_U\) and lifted across interface closure \(\widehat{A}_U^\star\) to super-tile boundary \(x_U(s)\).
        Lifting ascends along ancestor chains to root children \(C_s(r)\) and \(C_t(r)\).
        At the root (right panel), the left factor \(x_{C_s(r)}(s)\otimes J_{C_s(r)}^\top\) selects interface port \(p\in\Gamma_r\) reached from \(s\), the right factor \(J_{C_t(r)}\otimes y_{C_t(r)}(t)\) selects port \(q\in\Gamma_r\) reaching \(t\), and \(z_r(s,t)=1\) certifies path \(p\to q\) across \(\widehat{A}_r^\star\) (\(B_r=\varnothing\)).}
    \label{fig:query-propagation}
\end{figure}

A query reads one source row of \(R_{T_s}^{VB}\) and one target column of \(R_{T_t}^{BV}\) from the attachment relations \eqref{eq:base-attachment}. Those extracted vectors are written \(x_{T_s}(s)\in\mathbb{B}^{1\times b_{T_s}}\) and \(y_{T_t}(t)\in\mathbb{B}^{b_{T_t}\times 1}\),
\begin{align}
    \label{eq:query-base-vectors}
    (x_{T_s}(s))_p
     & =
    (R_{T_s}^{VB})_{sp},
     &
    (y_{T_t}(t))_q
     & =
    (R_{T_t}^{BV})_{qt},
\end{align}
so \((x_{T_s}(s))_p=1\) marks that boundary port \(p\in B_{T_s}\) is reachable from \(s\) inside \(T_s\) (Figure~\ref{fig:query-propagation}, Step~1), and \((y_{T_t}(t))_q=1\) marks that boundary port \(q\in B_{T_t}\) reaches \(t\) inside \(T_t\).
Since a super-tile does not keep an attachment from an interior cell to its ports, the same row and column are propagated up each ancestor chain.
For any non-root ancestor region \(W\) through which the frontier is propagated, let \(C_s(W)\) denote the unique child of \(W\) containing \(s\); define \(C_t(W)\) analogously on the target side.
A one-level propagation through \(W\) is then
\begin{align}
    \label{eq:query-propagation}
    x_W(s)
     & =
    \underbrace{x_{C_s(W)}(s)}_{\text{child frontier}}
    \otimes
    \underbrace{\Bigl( J_{C_s(W)}^{\top} \otimes \widehat{A}_W^\star \Bigr)}_{\text{interface expansion}}
    \otimes
    \underbrace{(P_W^\Gamma)^{\top}}_{\text{exterior projection}},
    \\
    \label{eq:query-propagation-y}
    y_W(t)
     & =
    \underbrace{P_W^\Gamma}_{\text{exterior projection}}
    \otimes
    \underbrace{\Bigl( \widehat{A}_W^\star \otimes J_{C_t(W)} \Bigr)}_{\text{interface expansion}}
    \otimes
    \underbrace{y_{C_t(W)}(t)}_{\text{child frontier}},
\end{align}
Equations~\eqref{eq:query-propagation}--\eqref{eq:query-propagation-y} define the corresponding source and target three-step lifting pipelines: each embeds child ports into \(\Gamma_W\), expands them across \(\widehat{A}_W^\star\), and projects onto the external ports of \(W\) via \(P_W^\Gamma\) (Figure~\ref{fig:query-propagation}, Step~2).

As hierarchical lifting ascends along both ancestor chains, the source and target frontiers eventually reach a common internal ancestor \(U\) of \(T_s\) and \(T_t\).
At this shared ancestor, the propagated source row and target column meet directly in the Boolean scalar
\begin{align}
    \label{eq:ancestor-test}
    z_U(s,t)
     & =
    \underbrace{\Bigl( x_{C_s(U)}(s) \otimes J_{C_s(U)}^{\top} \Bigr)}_{\text{embedded source frontier}}
    \otimes
    \underbrace{\widehat{A}_U^\star}_{\text{interface closure}}
    \otimes
    \underbrace{\Bigl( J_{C_t(U)} \otimes y_{C_t(U)}(t) \Bigr)}_{\text{embedded target frontier}},
\end{align}
and \(z_U(s,t)=1\) means that there are interface ports \(p,q \in \Gamma_U\) such that \(s\) reaches \(p\) through the source-side descendants, \(p\) reaches \(q\) inside \(U\), and \(q\) reaches \(t\) through the target-side descendants.
The ancestor test evaluates reachability over the full interface \(\Gamma_U\) rather than projecting via \(P_U^\Gamma\), because a connecting path inside \(U\) may traverse internal seams without ever touching the external boundary \(B_U\); furthermore, at the root tile, the external boundary is entirely empty, \(B_r = \emptyset\).

\begin{figure}[!t]
    \centering
    \resizebox{\linewidth}{!}{%
        \input{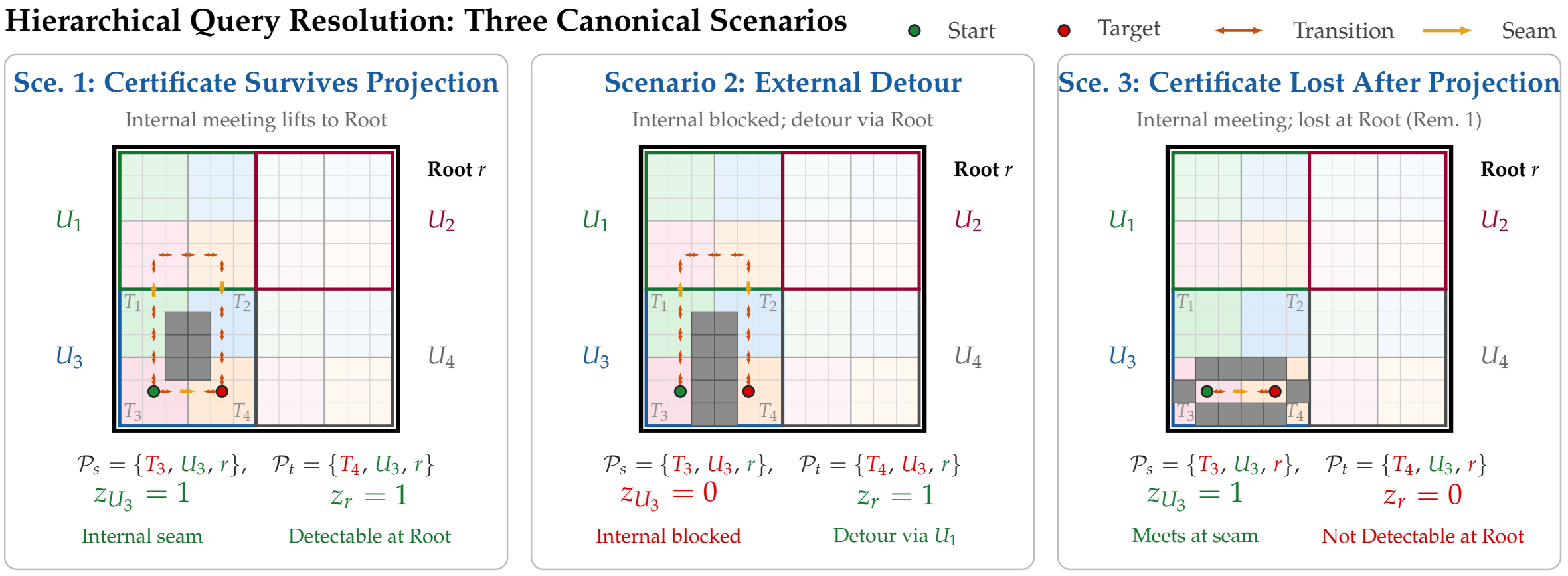}%
    }
    \caption{Hierarchical query resolution on three canonical topologies (\(b=3\), \(g=2\)).
        The source lies in base tile \(T_s=T_3\) and the target in \(T_t=T_4\).
        The query follows \(\mathcal{A}(T_s,T_t)=(U_3,r)\) upward from the lowest common ancestor \(U_3=\operatorname{LCA}(T_s,T_t)\), along \(\mathcal{P}_s\) and \(\mathcal{P}_t\).
        Root values for Scenarios~1 and 3 are shown counterfactually to illustrate the outcome if propagation continued beyond the positive \(U_3\) test.
        (\textit{Sce.~1}) Internal seams certify reachability immediately at \(U_3\) (\(z_{U_3}=1\), early exit), and the certificate survives projection to the root (\(z_r=1\)).
        (\textit{Sce.~2}) The internal seam at \(U_3\) is blocked (\(z_{U_3}=0\)); ascent along \(\mathcal{A}(T_s,T_t)\) to \(r\) finds an external detour through \(U_1\) (\(z_r=1\)).
        (\textit{Sce.~3}) The connection certificate at \(U_3\) (\(z_{U_3}=1\)) is lost under external-boundary projection \(P_{U_3}^\Gamma\) when lifting toward the root (\(z_r=0\)); early exit at the LCA prevents a false negative.}
    \label{fig:query-detour}
\end{figure}

A single common ancestor is not enough, since a directed path may leave a lower common ancestor and return through a larger enclosing region, as illustrated by the external detour in Figure~\ref{fig:query-detour} (Scenario~2).
Conversely, evaluating connection tests sequentially along the common ancestor sequence \(\mathcal{A}(T_s,T_t)\) starting from the lowest common ancestor \(\operatorname{LCA}(T_s,T_t)\) is essential: an internal path certified at the lowest common ancestor may fail to survive boundary projection to higher levels (Figure~\ref{fig:query-detour}, Scenario~3; Remark~\ref{rem:non-monotonicity}), whereas Scenario~1 illustrates a case in which the certificate remains valid at the root after projection (Figure~\ref{fig:query-detour}, Scenario~1).
If \(z_U(s,t)=1\), a connecting path is certified inside \(U\) and the query terminates immediately with reachability established (early termination).
Conversely, if \(z_U(s,t)=0\), the absence of a path inside \(U\) does not preclude an external detour through higher levels; lifting therefore continues along the common ancestor sequence \(\mathcal{A}(T_s,T_t)\) using~\eqref{eq:query-propagation}--\eqref{eq:query-propagation-y} until an ancestor test~\eqref{eq:ancestor-test} succeeds or the sequence is exhausted.

Write \(\mathcal{P}_s\) for the unique parent chain \(T_s\to\cdots\to r\) of \(T_s\), and \(\mathcal{P}_t\) likewise for \(T_t\).
Then \(\mathcal{A}(T_s,T_t)\) is the sequence of internal common ancestors ordered from the lowest common ancestor up to \(r\).
No member of \(\mathcal{A}(T_s,T_t)\) is a leaf, as \eqref{eq:ancestor-test} operates on internal interface closures.
When \(T_s=T_t\neq r\), a negative local test starts the common ancestor scan at the parent of the shared base tile.
Exhausting \(\mathcal{A}(T_s,T_t)\) without a positive test is conclusive, requiring no fallback search.
Algorithm~\ref{alg:query} states the resulting query, combining base extraction and hierarchical lifting (Figure~\ref{fig:query-propagation}) with the sequence of ancestor tests.

\begin{algorithm2e}[t]
    \caption{Exact Cell-to-Cell Reachability Query}
    \label{alg:query}
    \KwIn{Source \(s\), target \(t\), rooted H-BRICK index \(\mathcal{I}(G,\mathcal{T})\)}
    \KwOut{True if and only if \(t\) is reachable from \(s\)}
    Locate base tiles \(T_s\) and \(T_t\)\;
    \If{\(T_s=T_t\) and \((A_{T_s}^\star)_{st}=1\)}{\Return{true}}
    \If{\(T_s=T_t=r\)}{\Return{false}}
    Initialize \(x_{T_s}(s)\) from row \(s\) of \(R_{T_s}^{VB}\)\;
    Initialize \(y_{T_t}(t)\) from column \(t\) of \(R_{T_t}^{BV}\)\;
    Let \(\mathcal{A}(T_s,T_t)\) be the common internal ancestors, lowest first, ending at the root\;
    Let \(U_0\) be the first member of \(\mathcal{A}(T_s,T_t)\)\;
    Propagate \(x\) from \(T_s\) to \(C_s(U_0)\) and \(y\) from \(T_t\) to \(C_t(U_0)\) using \eqref{eq:query-propagation}--\eqref{eq:query-propagation-y}\;
    \For{each \(U \in \mathcal{A}(T_s,T_t)\)}{
        Compute \(z_U(s,t)\) using \eqref{eq:ancestor-test}\;
        \If{\(z_U(s,t)=1\)}{\Return{true}}
        \If{\(U\neq r\)}{
            Propagate \(x\) and \(y\) through \(U\) using \eqref{eq:query-propagation}--\eqref{eq:query-propagation-y}\;
        }
    }
    \Return{false}
\end{algorithm2e}

\begin{lemma}[Exactness of external-frontier propagation]
    \label{lem:propagation}
    Let \(C\) be a region on the source ancestor chain \(\mathcal{P}_s\) whose parent \(W=\operatorname{parent}(C)\) is non-root.
    Suppose that for every boundary port \(a\in B_C\), \((x_C(s))_a=1\) if and only if \(a\) is reachable from \(s\) by a directed path contained entirely in \(C\).
    Then the update \eqref{eq:query-propagation} produces, for every boundary port \(p\in B_W\),
    \begin{equation}
        \label{eq:frontier-support-x}
        (x_W(s))_p = 1
        \Longleftrightarrow
        p\text{ is reachable from }s\text{ by a directed path contained entirely in }W.
    \end{equation}
    Symmetrically, if \(C\) lies on the target ancestor chain \(\mathcal{P}_t\) with non-root parent \(W=\operatorname{parent}(C)\), and for every boundary port \(b\in B_C\), \((y_C(t))_b=1\) if and only if \(t\) is reachable from \(b\) by a directed path contained entirely in \(C\), then the update \eqref{eq:query-propagation-y} produces, for every boundary port \(q\in B_W\),
    \begin{equation}
        \label{eq:frontier-support-y}
        (y_W(t))_q = 1
        \Longleftrightarrow
        t\text{ is reachable from }q\text{ by a directed path contained entirely in }W.
    \end{equation}
    Thus, if the frontier is exact on the external boundary of \(C\), the updates \eqref{eq:query-propagation}--\eqref{eq:query-propagation-y} produce an exact frontier on the external boundary of \(W\).
\end{lemma}

\begin{proof}
    We prove the source-side equivalence \eqref{eq:frontier-support-x}; the target-side equivalence \eqref{eq:frontier-support-y} follows symmetrically on the reversed graph.
    Fix \(p\in B_W\).
    We prove the two implications separately.

    \noindent\textit{Forward implication.}
    Suppose first that \((x_W(s))_p=1\).
    Expanding \eqref{eq:query-propagation} entrywise, there must exist some \(a\in B_C\) such that \((x_C(s))_a=1\) and \((\widehat{A}_W^\star)_{\iota_C(a)\,p}=1\), where \(\iota_C(a)\in\Gamma_W\) is the unique interface index with \((J_C)_{\iota_C(a),a}=1\).
    By the hypothesis of the lemma, \((x_C(s))_a=1\) means that there exists a directed path from \(s\) to \(a\) contained entirely in \(C\).
    By Theorem~\ref{thm:composition}, the second relation means that there exists a directed path from \(a\) to \(p\) contained in \(W\).
    Concatenating these two paths gives a directed walk from \(s\) to \(p\) contained in \(W\), and hence \(p\) is reachable from \(s\) within \(W\).

    \noindent\textit{Reverse implication.}
    Now suppose that \(p\) is reachable from \(s\) by a directed path \(\pi\) contained entirely in \(W\).
    To show that \((x_W(s))_p=1\), it is sufficient, by \eqref{eq:query-propagation}, to identify a boundary port \(a\in B_C\) such that \((x_C(s))_a=1\) and \((\widehat{A}_W^\star)_{\iota_C(a)\,p}=1\).
    Choose \(a\) as the last vertex of the maximal prefix of \(\pi\) that remains inside \(C\).
    There are two cases.
    If \(\pi\) never leaves \(C\), then \(p\in C\).
    Since \(p\in B_W\), by \eqref{eq:port-set} the port \(p\) is incident to an edge whose other end lies outside \(W\); because \(C\subseteq W\), that end also lies outside \(C\).
    Hence \(p\in B_C\), and we take \(a=p\).
    Otherwise, the edge immediately following \(a\) leaves \(C\), so \(a\in B_C\) by \eqref{eq:port-set}.
    In either case, the prefix of \(\pi\) from \(s\) to \(a\) lies entirely in \(C\).
    Hence, by the hypothesis, \((x_C(s))_a=1\).
    The remaining suffix of \(\pi\) from \(a\) to \(p\) lies entirely in \(W\).
    Since \(a\in B_C\subseteq\Gamma_W\) and \(p\in B_W\subseteq\Gamma_W\), Theorem~\ref{thm:composition} implies from this suffix that \((\widehat{A}_W^\star)_{\iota_C(a)\,p}=1\), with the case \(a=p\) covered by reflexivity.
    Therefore the Boolean product in \eqref{eq:query-propagation} contains a witness \(a\) for which both factors are \(1\), and thus \((x_W(s))_p=1\).
    This establishes \eqref{eq:frontier-support-x}.
    Repeated application along \(\mathcal{P}_s\) and \(\mathcal{P}_t\), initialized at the base tiles by Lemma~\ref{lem:attachment}, therefore maintains this frontier-exactness invariant at every propagated level.
\end{proof}

\begin{remark}[Non-monotonicity of ancestor connection tests]
    \label{rem:non-monotonicity}
    Lemma~\ref{lem:propagation} characterizes only the reachability information exposed on the external boundary \(B_W\) of the current region; it does not imply that \(z_U(s,t)\) is monotone under ancestor inclusion.
    As illustrated by Scenario~3 in Figure~\ref{fig:query-detour}, a connection certified at \(U\) may rely on interface ports in \(\Gamma_U\setminus B_U\) without inducing compatible source-to-\(B_U\) and \(B_U\)-to-target frontiers.
    Such interface-only information is intentionally discarded by the external-boundary projection \(P_U^\Gamma\) during lifting; hence in general, \(z_U(s,t)=1 \centernot\Rightarrow z_W(s,t)=1\) for a strict ancestor \(W\supset U\).
    Consequently, Algorithm~\ref{alg:query} must evaluate \(z_U(s,t)\) at each common ancestor before propagating frontiers upward, and terminate immediately once \(z_U(s,t)=1\).
\end{remark}

\begin{theorem}[Query exactness]
    \label{thm:query}
    For any vertices \(s,t\in V\), Algorithm~\ref{alg:query} returns true if and only if \(t\) is reachable from \(s\) in \(G\).
\end{theorem}

\begin{proof}
    Suppose first that the procedure returns true.
    If \(s\) and \(t\) share a base tile and the local test \((A_{T_s}^\star)_{st}=1\) succeeds, then by definition of the local closure a directed path from \(s\) to \(t\) lies inside that tile, so \(t\) is reachable from \(s\) in \(G\).
    Otherwise the procedure returns true at a common internal ancestor \(U\) with \(z_U(s,t)=1\).
    By \eqref{eq:ancestor-test} there are interface ports \(p,q\in\Gamma_U\) such that \(p\) is reachable from \(s\) through the source-side descendants, \(q\) reaches \(t\) through the target-side descendants, and \(q\) is reachable from \(p\) inside \(U\); the first two statements follow from Lemma~\ref{lem:attachment} at the base level together with repeated application of Lemma~\ref{lem:propagation}, while the third follows from Theorem~\ref{thm:composition} applied to \(U\).
    Concatenating the three directed paths gives a directed walk from \(s\) to \(t\) in \(G\), hence a directed path, so every positive answer is sound.

    Conversely, suppose \(t\) is reachable from \(s\), with base tiles \(T_s\) and \(T_t\), and fix a directed path \(\pi\) from \(s\) to \(t\).
    If \(\pi\) lies entirely inside \(T_s=T_t\), the local test succeeds and the procedure returns true (including a one-tile hierarchy).
    Otherwise, let \(U\) be the lowest node of \(\mathcal{T}\) containing every vertex of \(\pi\).
    Then \(U\) is an internal common ancestor of \(T_s\) and \(T_t\), hence a member of \(\mathcal{A}(T_s,T_t)\).
    By minimality of \(U\), the fixed witness path \(\pi\) is not contained in any strict lower common ancestor \(U'\subset U\).
    If the scan nevertheless finds \(z_{U'}(s,t)=1\), Algorithm~\ref{alg:query} terminates early and correctly, possibly certifying a different \(s\)-to-\(t\) path.
    Otherwise, the fixed path \(\pi\) leaves \(U'\) and later re-enters it.
    The base attachments of Lemma~\ref{lem:attachment} together with repeated application of Lemma~\ref{lem:propagation} yield the exact external frontiers needed to continue the scan toward \(U\).
    When the scan reaches \(U\), vertices \(s\) and \(t\) belong to unique children \(C_s(U)\) and \(C_t(U)\) of \(U\).
    By minimality, \(\pi\) cannot lie entirely inside a single child of \(U\).
    Hence \(\pi\) leaves \(C_s(U)\) and eventually enters \(C_t(U)\); if \(C_s(U)=C_t(U)\), it necessarily leaves and later re-enters that child.
    Cutting \(\pi\) after its maximal prefix inside \(C_s(U)\) and before its maximal suffix inside \(C_t(U)\) produces ports \(p\in B_{C_s(U)}\) and \(q\in B_{C_t(U)}\) such that directed paths from \(s\) to \(p\) and from \(q\) to \(t\) lie entirely in \(C_s(U)\) and \(C_t(U)\) respectively, and a directed path from \(p\) to \(q\) lies in \(U\).
    By Lemma~\ref{lem:attachment} at the base level and Lemma~\ref{lem:propagation} under any subsequent propagation, \((x_{C_s(U)}(s))_p=1\) and \((y_{C_t(U)}(t))_q=1\), while Theorem~\ref{thm:composition} gives the corresponding entry of \(\widehat{A}_U^\star\), so \(z_U(s,t)=1\) by \eqref{eq:ancestor-test} and the procedure returns true.
\end{proof}

\section{Complexity Analysis}
\label{sec:analysis}

\begin{table}[t]
    \caption{Asymptotic preprocessing, persistent index storage, and query complexity on bounded-degree grid graphs. We assume \(M=O(N)\), \(K=\Theta(N)\), and fixed H-BRICK parameters \(b\) and \(g\); \(C\) and \(M_{\mathrm{cond}}\) denote the number of SCCs and condensation edges, respectively. TC denotes transitive closure.}
    \label{tab:complexity-comparison}
    \label{tab:baselines}
    \centering
    \small
    \setlength{\tabcolsep}{4pt}
    \renewcommand{\arraystretch}{1.15}
    \begin{tabularx}{\textwidth}{>{\raggedright\arraybackslash}p{0.24\textwidth} >{\raggedright\arraybackslash}X >{\raggedright\arraybackslash}X >{\raggedright\arraybackslash}X}
        \hline
        Method                                     & Preprocess                      & Index storage  & Query                                                    \\
        \hline
        BFS                                        & ---                             & \(O(1)\)       & \(O(N)\)                                                 \\
        SCC-DAG search                             & \(O(N)\)                        & \(O(N)\)       & \(O(C+M_{\mathrm{cond}})\)                               \\
        SCC-DAG closure                            & \(O(N)+C_{\mathrm{cl}}(C)\)     & \(O(N+C^2)\)   & \(O(1)\)                                                 \\
        GRAIL \cite{yildirim2010grail}             & \(O(N+k(C+M_{\mathrm{cond}}))\) & \(O(kC)\)      & \(O(k)\) to \(O(k+C+M_{\mathrm{cond}})\)\({}^{\dagger}\) \\
        O'Reach \cite{hanauer2021oreach}           & \(O(N)\)                        & \(O(N)\)       & \(O(1)\) to \(O(N)\)\({}^{\dagger}\)                     \\
        All-hub 2-hop \cite{cohen2003reachability} & \(O(N^2)\)                      & \(O(N^2)\)     & \(O(N)\)                                                 \\
        TC Warshall                                & \(O(N^3)\)                      & \(O(N^2)\)     & \(O(1)\)                                                 \\
        TC repeated squaring                       & \(O(N^3\log N)\)                & \(O(N^2)\)     & \(O(1)\)                                                 \\
        H-BRICK (fixed \(b,g\))                    & \(O(N^{3/2}\log N)\)            & \(O(N\log N)\) & \(O(N)\)                                                 \\
        \hline
    \end{tabularx}
    \vspace{2pt}
    \begin{minipage}{\textwidth}
        \footnotesize
        \textsuperscript{\dag} For GRAIL and O'Reach, the two query bounds denote index-only resolution and worst-case search fallback, respectively.
    \end{minipage}
\end{table}

This section analyzes storage, preprocessing, and query cost for the rooted index \(\mathcal{T}\) of Section~\ref{sec:hierarchy}.
Let \(C_{\mathrm{cl}}(d)\) denote the complexity of computing the reflexive transitive closure of a \(d\times d\) Boolean relation, and let \(C_{\mathrm{hyb}}(d)\) denote the cost of the adaptive bit-parallel closure implementation deployed in the benchmark.
The adaptive implementation chooses among the two alternatives based on relation dimension against a dimension threshold \(\theta_{\mathrm{hyb}}\):
relations of dimension \(d\le \theta_{\mathrm{hyb}}\) are closed by a word-parallel Warshall kernel~\cite{warshall1962theorem} requiring \(O(d^3/w)\) bitwise operations with machine word length \(w\);
relations of dimension \(d > \theta_{\mathrm{hyb}}\) are evaluated by bit-parallel repeated squaring~\eqref{eq:kleene-closure} using the Method of Four Russians (M4RM) Boolean matrix multiplication~\cite{arlazarov1970economical,albrecht2010algorithm,aho1974design}.
Each M4RM product costs \(O(d^3/(kw) + 2^k d^2/(kw))\) bitwise operations~\cite{albrecht2010algorithm,aho1974design} where \(k\) is the block width and \(w\) is the machine word length.
By~\eqref{eq:kleene-closure}, repeated squaring of \(I\oplus A\) reaches the reflexive transitive closure in at most \(\lceil\log_2(d-1)\rceil\) matrix multiplications (Section~\ref{sec:preliminaries}).
Because both \(k\) and \(w\) are fixed implementation constants, the conservative worst-case bound of the adaptive realization is
\begin{align}
    \label{eq:closure-cost-hybrid}
    C_{\mathrm{hyb}}(d)
     & =
    O\left(
    \left[\frac{d^3}{kw} + \frac{2^k d^2}{kw}\right] \log d
    \right)
    =
    O\left(d^3\log d\right).
\end{align}

\subsection{Storage and Boundary Compression}

A base tile stores the local closure required by same-tile queries, its boundary summary, and its two attachment relations, which in dense form occupy \(O(n_T^2)\), \(O(b_T^2)\), and \(O(n_Tb_T)\) entries each, so \(M_T^{\mathrm{base}} = O(n_T^2 + b_T^2 + 2n_Tb_T) = O(n_T^2)\) by \(b_T\leq n_T\).
A super-tile stores the interface closure \(\widehat{A}_U^\star\), of dimension \(\gamma_U\times\gamma_U\), and, if it is not the root, the exposed summary \(S_U\), of dimension \(b_U\times b_U\), so \(M_U^{\mathrm{super}}=O(\gamma_U^2+b_U^2)=O(\gamma_U^2)\).
The total dense index storage is therefore
\begin{align}
    \label{eq:storage-total}
    M_{\mathrm{index}}
     & =
    O\left(
    \sum_{T\in\mathcal{T}^{(0)}}
    \left(n_T^2 + b_T^2 + 2n_Tb_T\right)
    +
    \sum_{\ell\geq 1}
    \sum_{U\in\mathcal{T}^{(\ell)}}
    \left(\gamma_U^2 + b_U^2\right)
    \right).
\end{align}

\begin{corollary}[Regular grid decomposition with uniform aggregation]
    \label{cor:grid-storage}
    Let the map occupy a square area of \(K\) cells, of which \(N=|V|\leq K\) are passable, and decompose it into base tiles of side \(b\).
    Let every super-tile group \(g\times g\) adjacent children for a uniform integer \(g\geq 2\), and for simplicity suppose \(\sqrt{K}/b=g^h\), so the hierarchy reaches its root after \(h=\log_g(\sqrt{K}/b)\) super-levels.
    Then \(n_T=O(b^2)\), \(b_T=O(b)\), and \(|\mathcal{T}^{(0)}|=\Theta(K/b^2)\), so the level-\(0\) terms of \eqref{eq:storage-total} satisfy
    \begin{align}
        \label{eq:grid-storage}
        \sum_{T\in\mathcal{T}^{(0)}} n_T^2
         & =
        O(Kb^2),
         &
        \sum_{T\in\mathcal{T}^{(0)}} n_Tb_T
         & =
        O(Kb),
         &
        \sum_{T\in\mathcal{T}^{(0)}} b_T^2
         & =
        O(K),
    \end{align}
    whereas a dense materialization of the full transitive closure requires \(\Theta(N^2)\) Boolean entries.
    The complete index satisfies
    \begin{align}
        \label{eq:grid-storage-total}
        M_{\mathrm{index}}
         & =
        O\left(Kb^2+Kg^2h\right)
        =
        O\left(Kb^2+Kg^2\log_g(\sqrt{K}/b)\right).
    \end{align}
\end{corollary}

\begin{proof}
    Multiply the per-tile bounds \(n_T^2=O(b^4)\), \(n_Tb_T=O(b^3)\), and \(b_T^2=O(b^2)\) by the tile count \(\Theta(K/b^2)\).
    At super-level \(\ell\in\{1,\ldots,h\}\), a region has side length \(L_\ell=bg^\ell\), and there are \(\Theta(K/L_\ell^2)=\Theta(K/(b^2g^{2\ell}))\) such regions.
    Each parent contains \(g^2\) children of side \(L_{\ell-1}\), so its interface and external boundary satisfy \(\gamma_U=O(g^2L_{\ell-1})=O(gL_\ell)\) and \(b_U=O(L_\ell)\), respectively.
    The storage at super-level \(\ell\) is therefore
    \begin{align*}
        \sum_{U\in\mathcal{T}^{(\ell)}}
        \left(\gamma_U^2+b_U^2\right)
         & =
        O\left(
        \frac{K}{L_\ell^2}
        \left(g^2L_\ell^2+L_\ell^2\right)
        \right)
        =
        O(Kg^2).
    \end{align*}
    Summing over the \(h\) super-levels and adding the level-\(0\) terms gives \eqref{eq:grid-storage-total}.
\end{proof}

Partial edge groups and bounded-aspect-ratio rectangular maps satisfy the same upper bounds up to constants.
For fixed \(g\), \eqref{eq:grid-storage-total} reduces to \(O(Kb^2+K\log(K/b^2))\); the factor \(g^2h\) shows that wider aggregation shortens the hierarchy but enlarges intermediate interfaces.
Larger base tiles raise \(C_{\mathrm{hyb}}(n_T)\) and base storage but reduce interface ports carried upward.
As a theoretical baseline for comparison, a flat single-level decomposition has \(P=\Theta(K/b)\) interface ports (generally \(P=O(K/b)\)) and requires \(\Theta(P^2)=\Theta(K^2/b^2)\) dense closure storage.
On that boundary graph, a conventional cubic closure kernel such as Warshall's algorithm~\cite{warshall1962theorem} (cf. Table~\ref{tab:complexity-comparison}) requires \(\Theta(P^3)=\Theta((K/b)^3)\) scalar work, whereas repeated squaring under dense Boolean multiplication requires \(O(P^3\log P)\), both of which H-BRICK's recursive hierarchy avoids.
The cell count \(K\) rather than \(N\) governs these sums because tiles are spatial regions.

\subsection{Preprocessing Complexity}
\label{sec:preprocessing-complexity}

For a base tile the dominant cost is forming the local closure \(A_T^\star\) via the adaptive closure engine, with cost \(C_{\mathrm{hyb}}(n_T)=O(n_T^3\log n_T)\); additionally, extracting the summary and the two attachment relations from that closure costs at most \(O(b_T^2+2n_Tb_T)\) in dense form.
For a super-tile the dominant cost is forming \(\widehat{A}_U^\star\) via the adaptive closure engine on the \(\gamma_U\times\gamma_U\) interface, with cost \(C_{\mathrm{hyb}}(\gamma_U)=O(\gamma_U^3\log\gamma_U)\).
Assembling the interface \(\widehat{A}_U\) from child summaries and one-step seam edges costs \(O(\gamma_U^2)\) to write the matrix with \(A_U^{\mathrm{seam}}\) and \(O\bigl(\sum_{T_i\in\operatorname{ch}(U)} b_{T_i}^2\bigr)\) to place each embedding \(J_i\otimes S_{T_i}\otimes J_i^{\top}\).
If \(U\neq r\), extracting the exposed boundary submatrix \(S_U=P_U^\Gamma\otimes\widehat{A}_U^\star\otimes(P_U^\Gamma)^{\top}\) via the \(0\)--\(1\) selection matrix \(P_U^\Gamma\) costs \(O(b_U^2)\).
Collecting the levels,
\begin{align}
    \label{eq:preprocessing-total}
    T_{\mathrm{pre}}
     & =
    O\left(
    \sum_{T\in\mathcal{T}^{(0)}}
    C_{\mathrm{hyb}}(n_T)
    +
    \sum_{\ell\geq 1}
    \sum_{U\in\mathcal{T}^{(\ell)}}
    C_{\mathrm{hyb}}(\gamma_U)
    \right)
    +
    T_{\mathrm{embed}},
\end{align}
where \(T_{\mathrm{embed}}\) is the lower-order cost of forming embedded interface graphs and applying boundary projections.

Using the conservative \(O(d^3\log d)\) worst-case bound of the adaptive closure engine and the regular decomposition of Corollary~\ref{cor:grid-storage}, the base level costs \(\Theta(K/b^2)\cdot O(b^6\log b)=O(Kb^4\log b)\).
At super-level \(\ell\), multiplying the \(\Theta(K/(b^2g^{2\ell}))\) regions by the closure cost for \(\gamma_U=O(bg^{\ell+1})\) gives \(T_{\mathrm{pre}}^{(\ell)}=O(Kb\,g^{\ell+3}\log(bg^{\ell+1}))\).
Summing these costs across all \(\ell=1,\ldots,h\) super-levels gives
\begin{align}
    \sum_{\ell=1}^{h} T_{\mathrm{pre}}^{(\ell)}
     & = O\left( Kb\,g^3 \log(g\sqrt{K}) \sum_{\ell=1}^h g^\ell \right)
    = O\left( Kb\,g^3 \log(g\sqrt{K}) \cdot \frac{g^{h+1}-g}{g-1} \right).
\end{align}
Since \(bg^h=\sqrt{K}\) and \(g/(g-1)\leq 2\) for all \(g\geq 2\), the summation satisfies \(b\cdot\frac{g^{h+1}-g}{g-1} = \frac{g}{g-1}(\sqrt{K}-b) = O(\sqrt{K})\), yielding the total super-level preprocessing cost \(O(K^{3/2}g^3\log(g\sqrt{K}))\).
Writing and projecting the interface matrices adds \(T_{\mathrm{embed}}=O(Kg^2h)\), which is lower order, hence
\begin{equation}
    \label{eq:grid-preprocessing}
    T_{\mathrm{pre}} = O\left(Kb^4\log b + K^{3/2}g^3\log(g\sqrt{K})\right),
\end{equation}
which simplifies to \(O(N^{3/2}\log N)\) for fixed \(b,g\).

\subsection{Query Complexity}
\label{sec:query-complexity}

A worst-case query lifts along both ancestor chains \(\mathcal{P}_s\) and \(\mathcal{P}_t\) and tests every common internal ancestor in \(\mathcal{A}(T_s,T_t)\).
Let \(\mathcal{P}_s^{\mathrm{prop}} = \{W \in \mathcal{P}_s \setminus \{r\} : W\text{ is internal}\}\) and \(\mathcal{P}_t^{\mathrm{prop}} = \{W \in \mathcal{P}_t \setminus \{r\} : W\text{ is internal}\}\) denote the intermediate regions executing upward frontier propagation~\eqref{eq:query-propagation}--\eqref{eq:query-propagation-y}.
Base boundary extraction costs \(O(b)\), while dense frontier propagation across \(\widehat{A}_W^\star\) or an ancestor test across \(\widehat{A}_U^\star\) costs \(O(\gamma_W^2)\) or \(O(\gamma_U^2)\), so
\begin{equation}
    \label{eq:query-cost}
    T_{\mathrm{query}} = O\left(
    \sum_{W\in\mathcal{P}_s^{\mathrm{prop}}} \gamma_W^2
    + \sum_{W\in\mathcal{P}_t^{\mathrm{prop}}} \gamma_W^2
    + \sum_{U\in\mathcal{A}(T_s,T_t)} \gamma_U^2
    \right).
\end{equation}
In the regular \(g\times g\) hierarchy, each chain contains at most one region per super-level, so \(\gamma_U=O(bg^{\ell+1})\) gives
\begin{equation}
    \label{eq:grid-query}
    T_{\mathrm{query}} = O\left(\sum_{\ell=1}^{h} b^2g^{2\ell+2}\right) = O(g^2K).
\end{equation}
For fixed \(g\), this remains \(O(K)\), so H-BRICK does not offer an asymptotic query-time improvement over BFS in the worst case.
However, because Algorithm~\ref{alg:query} terminates immediately at the lowest ancestor certifying reachability, a query settling at ancestor \(U_{s,t}\) of area \(K_{s,t}\) incurs only \(T_{\mathrm{query}}(s,t) = O(g^2 K_{s,t})\) (\(O(K_{s,t})\) for fixed \(g\)).
Under the \(w\)-bit word-RAM model, dense frontier propagation and ancestor tests require \(O(\gamma^2/w)\) bitwise word operations, scaling worst-case queries to \(O(g^2 K / w)\) word operations; Table~\ref{tab:complexity-comparison} summarizes these asymptotic bounds alongside baseline reachability methods on grid graphs.

\section{Implementation Optimizations}
\label{sec:variants}

The construction of Sections~\ref{sec:boundary-kleene}--\ref{sec:hierarchy} admits several implementation optimizations that reduce preprocessing and query cost without changing the stored reachability relations or their exactness guarantees.
This section summarizes the closure backend, the skip-lift realization, and the batched query algorithm used in the experiments.
The canonical reference (\texttt{HBrick}) provides the baseline by directly evaluating queries through the level-by-level recurrence~\eqref{eq:query-propagation} and ancestor tests~\eqref{eq:ancestor-test} over elementary parent--child interface summaries.

\subsection{Adaptive Bit-Parallel Closure Backend}
\label{subsec:closure-backend}

Closure computation is the main preprocessing operation at both base tiles and super-tile interfaces.
Rather than using a single closure kernel for all matrix sizes, the implementation selects between two bit-parallel methods according to relation dimension.
For \(d \le \theta_{\mathrm{hyb}}\) (\(\theta_{\mathrm{hyb}}=512\) in our implementation), closure is computed using a packed Warshall kernel.
Larger relations are evaluated by repeated squaring with M4RM Boolean matrix multiplication, which groups several Boolean row combinations into table lookups.
Both kernels compute the same reflexive transitive closure and therefore do not alter the algebraic construction of Sections~\ref{sec:boundary-kleene}--\ref{sec:hierarchy}.

\subsection{Direct Multi-Level Skip Lifts (\texttt{HBrickSkipLift})}
\label{subsec:skiplift}

The canonical \texttt{HBrick} realization evaluates cross-tile queries by extracting base-tile frontiers \(x_{T_s}(s) = (R_{T_s}^{VB})_{s, :}\) and \(y_{T_t}(t) = (R_{T_t}^{BV})_{:, t}\), ascending the hierarchy step by step through recurrence~\eqref{eq:query-propagation}, and testing common ancestors via~\eqref{eq:ancestor-test}.
To eliminate runtime pointer indirection and intermediate vector--matrix products during upward traversal, \texttt{HBrickSkipLift} precomputes composite boundary-to-boundary skip-lift operators during index construction.

For each parent--child tree edge with non-root parent \(W\) and child \(C\), the elementary single-step forward lift operator fusing child embedding, interface closure multiplication, and exterior boundary projection is:
\begin{align}
    \label{eq:step-lift-matrix}
    L^{\mathrm{fwd}}_{C\to W} & = J_C^{\top} \otimes \widehat{A}_W^\star \otimes (P_W^\Gamma)^{\top} \in \mathbb{B}^{b_C \times b_W};
\end{align}
the reverse lift \(L^{\mathrm{bwd}}_{W\leftarrow C} = P_W^\Gamma \otimes \widehat{A}_W^\star \otimes J_C \in \mathbb{B}^{b_W \times b_C}\) is defined analogously.
For a base tile \(T\) and any proper non-root ancestor \(W\) along the ascending ancestor chain \(T = W_0, W_1, \dots, W_\ell = W\), \texttt{HBrickSkipLift} precomputes and stores the composite boundary-to-boundary forward skip-lift matrix:
\begin{align}
    \label{eq:skip-lift-forward}
    \Lambda^{\mathrm{fwd}}_{T \to W} & = L^{\mathrm{fwd}}_{W_0\to W_1} \otimes L^{\mathrm{fwd}}_{W_1\to W_2} \otimes \cdots \otimes L^{\mathrm{fwd}}_{W_{\ell-1}\to W} \in \mathbb{B}^{b_T \times b_W},
\end{align}
and symmetrically the composite backward skip-lift matrix \(\Lambda^{\mathrm{bwd}}_{W \leftarrow T} = L^{\mathrm{bwd}}_{W\leftarrow W_{\ell-1}} \otimes \cdots \otimes L^{\mathrm{bwd}}_{W_1\leftarrow W_0} \in \mathbb{B}^{b_W \times b_T}\), adopting the identity convention \(\Lambda^{\mathrm{fwd}}_{T \to T} := I_{b_T}\) and \(\Lambda^{\mathrm{bwd}}_{T \leftarrow T} := I_{b_T}\).

During query execution, the base frontier is obtained by direct row lookup \(x_T(s) = (R_T^{VB})_{s, :} \in \mathbb{B}^{1 \times b_T}\) and propagated to any requested ancestor \(W\) in a single bit-parallel vector--matrix application:
\begin{align}
    \label{eq:skip-lift-eval}
    x_W(s) = x_T(s) \otimes \Lambda^{\mathrm{fwd}}_{T \to W} \in \mathbb{B}^{1 \times b_W}.
\end{align}
Symmetrically, the backward frontier is \(y_W(t) = \Lambda^{\mathrm{bwd}}_{W \leftarrow T} \otimes y_T(t) \in \mathbb{B}^{b_W \times 1}\) with \(y_T(t) = (R_T^{BV})_{:, t}\).
At each common ancestor \(U\), arriving frontiers are embedded into \(\Gamma_U\) via \(x_{C_s(U)}(s) \otimes J_{C_s(U)}^\top\) and \(J_{C_t(U)} \otimes y_{C_t(U)}(t)\), and the reachability meet is evaluated across \(\widehat{A}_U^\star\) as in~\eqref{eq:ancestor-test}.
By associativity of Boolean matrix multiplication, \eqref{eq:skip-lift-eval} produces exactly the same frontier as sequential application of \eqref{eq:query-propagation}; the backward direction is analogous.
Hence \texttt{HBrickSkipLift} changes only the evaluation schedule, not the reachability result.
The precomputed operators enlarge the index beyond the bound of Corollary~\ref{cor:grid-storage}: under the assumptions of that corollary they occupy \(O(K^{3/2}/(bg))\) bits in total, which is \(O(K^{3/2})\) for fixed \(b\) and \(g\) and thus asymptotically dominates the \(O(Kb^2 + Kg^2 h)\) base index.
The measured storage--latency trade-off is reported in Section~\ref{sec:results}.

\subsection{Batched Many-Source/Many-Target Queries}
\label{subsec:batch-hbrick}

In multi-agent task allocation, fleet coordination, and frontier exploration, reachability queries frequently arrive in batches over common vertex sets.
Given a source set \(\mathcal{S} = \{s_1, \dots, s_m\}\) and a destination set \(\mathcal{D} = \{t_1, \dots, t_n\}\), the batched query evaluates the full \(m \times n\) Boolean reachability matrix \(Q \in \mathbb{B}^{m \times n}\) with \(Q[i, j] = \operatorname{Reach}_G(s_i, t_j)\).
Rather than evaluating \(m \times n\) independent scalar queries through the hierarchy, the batched query algorithm amortizes hierarchical traversal by sharing intermediate boundary lifts and interface reachability frontiers across eligible query pairs.

For pairs residing in the same base tile (\(T_{s_i} = T_{t_j}\)), a positive local closure entry \((A_{T_{s_i}}^\star)_{s_i, t_j} = 1\) resolves reachability immediately; a negative local entry is conclusive only when that base tile is the root (\(T_{s_i} = r\)).
The ancestor chains of the endpoints determine the first hierarchy level at which each source--destination pair shares a common ancestor; the pair is not tested below that level.
The batched query is then evaluated bottom-up, one hierarchy level at a time.
At a fixed level, each endpoint belongs to one ancestor node \(U\) on its chain, and computations are shared among pairs whose chains coincide at the same \(U\).
At each level:
(1)~each source \(s_i\) with a valid frontier is lifted to the required child boundary of \(U\), using the precomputed skip lift of Section~\ref{subsec:skiplift} when needed, and its source-side interface reachability frontier \(\phi_{s_i, U} = x_{C_s(U)}(s_i) \otimes J_{C_s(U)}^\top \otimes \widehat{A}_U^\star \in \mathbb{B}^{1 \times \gamma_U}\) is computed once;
(2)~each destination frontier is similarly lifted and embedded into \(\Gamma_U\) as \(J_{C_t(U)} \otimes y_{C_t(U)}(t_j) \in \mathbb{B}^{\gamma_U \times 1}\) once; and
(3)~every unresolved pair eligible at \(U\) is tested by:
\begin{align}
    \label{eq:batch-meet-eval}
    \phi_{s_i, U} \otimes J_{C_t(U)} \otimes y_{C_t(U)}(t_j) = 1.
\end{align}
In the packed-bit implementation, this scalar Boolean product is evaluated as a nonempty bitwise intersection between the source-side interface frontier and the embedded destination frontier.
A positive test resolves the corresponding pair immediately, whereas a negative test remains inconclusive and the pair is retained for evaluation at higher common ancestors up to the root.
The prepared destination frontier is reused across all unresolved sources tested against that destination at the same ancestor \(U\), and the source-side frontier \(\phi_{s_i, U}\) is reused symmetrically across eligible destinations.
By associativity of Boolean matrix multiplication, \eqref{eq:batch-meet-eval} is a reassociation of the ancestor test \(z_U(s_i, t_j)\) of~\eqref{eq:ancestor-test}, with the interface closure \(\widehat{A}_U^\star\) applied on the source side once and shared across eligible destinations.
Pairs failing this test across all shared ancestors are resolved as unreachable.

\begin{proposition}[Batch-query exactness]
    \label{prop:batch-exact}
    For every finite source set \(\mathcal{S}=\{s_1,\dots,s_m\}\subseteq V\) and destination set \(\mathcal{D}=\{t_1,\dots,t_n\}\subseteq V\), the batched procedure of Section~\ref{subsec:batch-hbrick} returns a matrix \(Q\in\mathbb{B}^{m\times n}\) with \(Q[i,j]=\operatorname{Reach}_G(s_i,t_j)\) for every \(s_i\in \mathcal{S}\) and every \(t_j\in \mathcal{D}\).
\end{proposition}

\begin{proof}
    Same-tile pairs use the same local closure test as Algorithm~\ref{alg:query}.
    For every remaining pair \((s_i, t_j)\), the batched procedure evaluates the same common-ancestor tests as Algorithm~\ref{alg:query}, differing only by sharing endpoint lifts and source-side frontiers \(\phi_{s_i, U}\) across eligible pairs at each ancestor node.
    By associativity of Boolean matrix multiplication, the meet test~\eqref{eq:batch-meet-eval} is identical to the scalar ancestor certificate \(z_U(s_i, t_j)\) of~\eqref{eq:ancestor-test}.
    Every batched decision therefore coincides with the scalar query pair by pair, and exactness follows from Theorem~\ref{thm:query}.
\end{proof}

\section{Computational Evaluation and Results}
\label{sec:experimental-setup}
\label{sec:results}

\begin{table}[!t]
    \caption{Evaluation maps from the Moving AI repository \cite{sturtevant2012benchmarks}, native dimensions, passable vertex count \(N\), geometric edge count \(M_{\mathrm{geom}}\), obstacle fraction, bidirectional range \(p_{\mathrm{bi}}\), and the storage required by a dense bit-packed full transitive closure (\(\lceil N^2/8\rceil\) bytes).}
    \label{tab:benchmark-graphs}
    \begin{adjustwidth}{-\extralength}{0cm}
        {\raggedleft
\begingroup
\setlength{\fboxsep}{0pt}%
\setlength{\fboxrule}{0.25pt}%
\setlength{\tabcolsep}{3.0pt}%
\newcommand{\mapimg}[1]{%
    \fcolorbox{black!40}{white}{\includegraphics[width=0.22\textwidth,interpolate=false]{figures/thumbs/#1.png}}}%
\resizebox{\textwidth}{!}{%
\begin{tabular}{@{}cccc@{}}
\mapimg{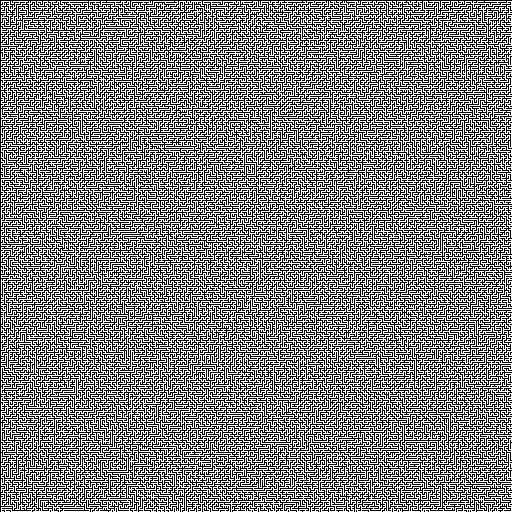} & \mapimg{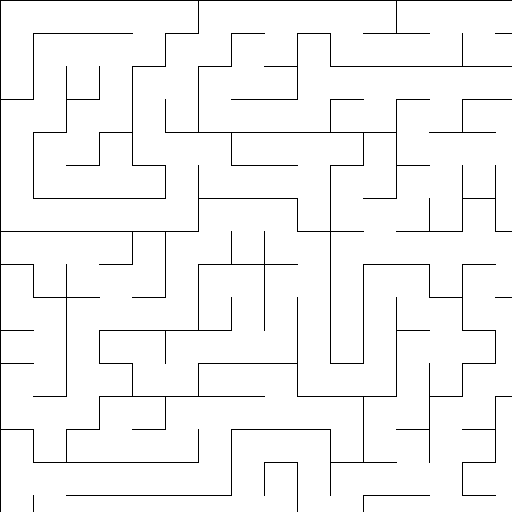} & \mapimg{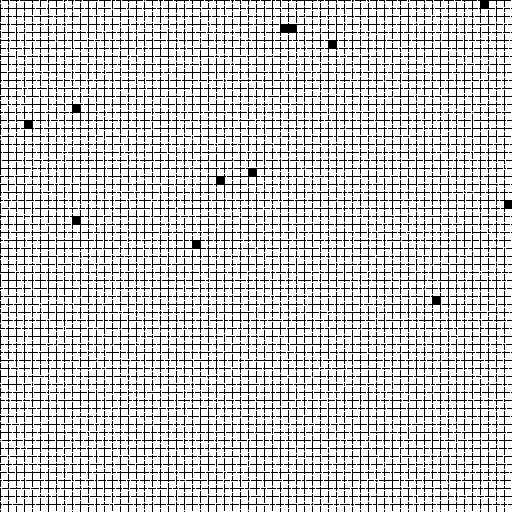} & \mapimg{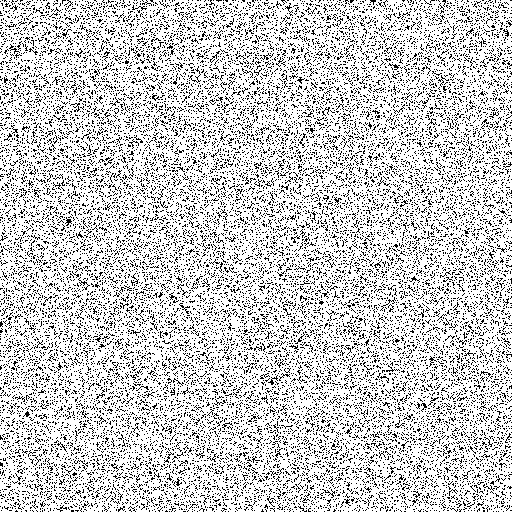} \\[-1.5pt]
{\scriptsize\Map{1}} &
{\scriptsize\Map{2}} &
{\scriptsize\Map{3}} &
{\scriptsize\Map{4}} \\[3.5pt]
\mapimg{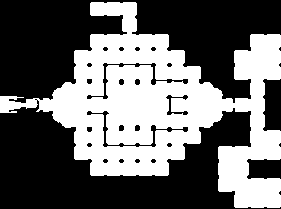} & \mapimg{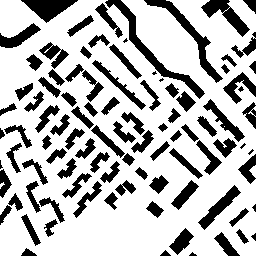} & \mapimg{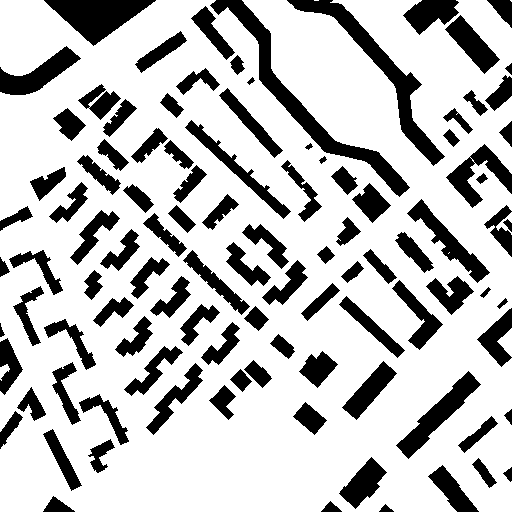} & \mapimg{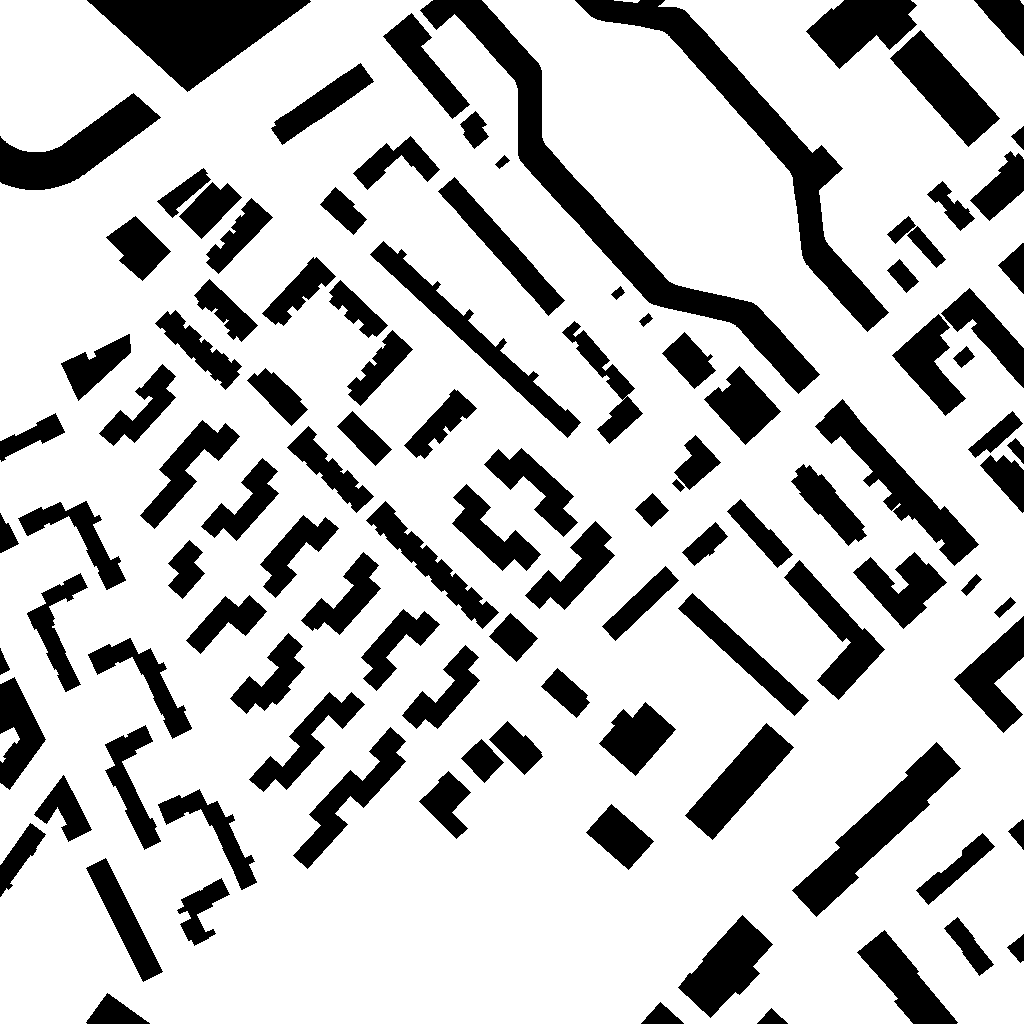} \\[-1.5pt]
{\scriptsize\Map{5}} &
{\scriptsize\Map{6}} &
{\scriptsize\Map{7}} &
{\scriptsize\Map{8}}
\end{tabular}}%
\endgroup
\vspace{3.5pt}
\par}
        \centering
        \footnotesize
        \setlength{\tabcolsep}{4.0pt}
        \renewcommand{\arraystretch}{1.12}
        \begin{tabular*}{\fulllength}{@{\extracolsep{\fill}}c l l c r r c c c r@{}}
            \hline
            Map     & Name                     & Structural Domain       & Dimensions           & \(N\)   & \(M_{\mathrm{geom}}\) & \(M_{\mathrm{geom}}/N\) & Obst.  & \(p_{\mathrm{bi}}\) range & TC (GiB) \\
            \hline
            \Map{1} & \texttt{maze512-1-0}     & 1-corridor maze         & \(512 \times 512\)   & 131,071 & 131,070               & 1.00                    & 50.0\% & [0.20, 1.00]              & 2.000    \\
            \Map{2} & \texttt{maze512-32-0}    & 32-corridor maze        & \(512 \times 512\)   & 253,840 & 499,377               & 1.97                    & 3.2\%  & [0.06, 0.31]              & 7.501    \\
            \Map{3} & \texttt{8room\_000}      & 8-room building layout  & \(512 \times 512\)   & 206,642 & 356,006               & 1.72                    & 21.2\% & [0.57, 0.85]              & 4.971    \\
            \Map{4} & \texttt{random512-20-0}  & 20\% random obstacles   & \(512 \times 512\)   & 209,281 & 334,908               & 1.60                    & 20.2\% & [0.32, 0.61]              & 5.099    \\
            \Map{5} & \texttt{arena2}          & Game arena (Dragon Age) & \(281 \times 209\)   & 24,311  & 47,085                & 1.94                    & 58.6\% & [0.04, 0.53]              & 0.069    \\
            \Map{6} & \texttt{Boston\_0\_256}  & Small street network          & \(256 \times 256\)   & 47,768  & 90,649                & 1.90                    & 27.1\% & [0.05, 0.43]              & 0.266    \\
            \Map{7} & \texttt{Boston\_0\_512}  & Medium street network   & \(512 \times 512\)   & 196,725 & 383,664               & 1.95                    & 25.0\% & [0.04, 0.33]              & 4.505    \\
            \Map{8} & \texttt{Boston\_0\_1024} & Large city network      & \(1024 \times 1024\) & 796,896 & 1,574,247             & 1.98                    & 24.0\% & [0.02, 0.35]              & 73.929   \\
            \hline
        \end{tabular*}
    \end{adjustwidth}
\end{table}

This section evaluates the computational performance of the boundary-compressed Kleene hierarchy against representative reachability baselines: unindexed breadth-first search (\texttt{BFS}), strongly connected component (SCC) condensation followed by search on the condensation DAG (\texttt{SCC-DAG}) \cite{tarjan1972depth}, and the interval-labeling indexes GRAIL \cite{yildirim2010grail} and O'Reach \cite{hanauer2021oreach}.
Exact 2-hop labeling \cite{cohen2003reachability} is also a competitive exact index.
However, it completed only on 11 of the 48 instances and was skipped by the memory policy on the other 37, so Table~\ref{tab:external-regime-summary} reports its coverage rather than aggregate latencies, and it is otherwise compared through the asymptotic bounds of Table~\ref{tab:complexity-comparison}.
We quantify index memory footprint, single-pair query latency across directional flow regimes, and batched evaluation throughput.
All comparisons in this section use the same 48 instances and the same query pairs.

\subsection{Datasets and Measurement Protocol}
\label{subsec:datasets}
\label{subsec:directed-construction}
\label{subsec:protocol}

The benchmark suite comprises eight public 2D occupancy grids from the Moving AI repository \cite{sturtevant2012benchmarks}, spanning city street networks, game arenas, mazes, room layouts, and random obstacle fields (Table~\ref{tab:benchmark-graphs}).
While these grid graphs are natively undirected 4-neighbor planar lattices, directed instances are constructed under two complementary orientation models:
(1)~\emph{random asymmetric} (\(V01\)--\(V05\)), where adjacent passable cells are connected bidirectionally with probability \(p_{\mathrm{bi}}\) and unidirectionally with probability \(1-p_{\mathrm{bi}}\), inducing a directed percolation phase transition; and
(2)~\emph{gradient flow} (\(V06\)), where edges are directed at a flow angle of \(\theta = 45^\circ\) with a backflow probability \(p_{\mathrm{back}} = 0.05\) (\(p_{\mathrm{bi}} = 0\)).

Let \(V_{\mathrm{pass}}\) denote the set of passable grid vertices with \(N = |V_{\mathrm{pass}}|\). We formally define reachability density as
\begin{equation}
    \label{eq:reachability-density}
    \rho = \frac{|\{(s, t) \in V_{\mathrm{pass}} \times V_{\mathrm{pass}} : s \neq t,\ \operatorname{Reach}_G(s,t) = 1\}|}{N(N - 1)},
\end{equation}
To systematically evaluate performance across connectivity levels, each benchmark map is calibrated so that \(\rho\) follows a geometrically spaced progression from \(\rho \approx 0.95\) in regime \(V01\) down to \(\rho \approx 0.03\) in regime \(V05\).
Because obstacle topology and doorway bottlenecks vary significantly across layouts, a uniform \(p_{\mathrm{bi}}\) yields vastly disparate reachability; \(p_{\mathrm{bi}}\) is therefore calibrated per map via binary search over \([0, 1]\) until empirical reachability matches target density \(\rho\)~\eqref{eq:reachability-density} (Table~\ref{tab:calibration-ladder}), yielding \(48\) calibrated benchmark instances (\(8\) maps \(\times\) \(6\) regimes).
\Map{1} represents an exceptional topological regime: because it is an acyclic unit-width corridor maze lacking lattice cycles, random edge orientation fragments the underlying tree into isolated directed paths rather than forming a percolation giant component; its calibration is therefore uniquely governed by strongly connected component partitioning following \(|\mathrm{SCC}| \approx (1 - p_{\mathrm{bi}})N\).

\begin{table}[b]
    \caption{Calibrated reachability density \(\rho\) and bidirectional edge probability \(p_{\mathrm{bi}}\) across orientation regimes \(V01\)--\(V06\) for the eight benchmark maps. On \Map{1}, the tree-like maze is calibrated via strongly connected component (SCC) partitioning following \(|\mathrm{SCC}| \approx (1 - p_{\mathrm{bi}})N\).}
    \label{tab:calibration-ladder}
    \centering
    \scriptsize
    \setlength{\tabcolsep}{3.5pt}
    \renewcommand{\arraystretch}{1.12}
    \begin{tabular*}{\linewidth}{@{\extracolsep{\fill}}l c c c c c c@{}}
        \toprule
        Map     & \(V01\) (\(\approx 95\%\)) & \(V02\) (\(\approx 70\%\)) & \(V03\) (\(\approx 50\%\)) & \(V04\) (\(\approx 16\%\)) & \(V05\) (\(\approx 3\%\)) & \(V06\) (Gradient) \\
        \midrule
        \Map{1} & 100\% (\(1.00\))           & 20\% SCC (\(0.80\))        & 40\% SCC (\(0.60\))        & 60\% SCC (\(0.40\))        & 80\% SCC (\(0.20\))       & DAG (\(0.00\))     \\
        \Map{2} & 95.6\% (\(0.31\))          & 62.6\% (\(0.18\))          & 27.8\% (\(0.16\))          & 6.8\% (\(0.09\))           & 3.1\% (\(0.06\))          & 1.0\%              \\
        \Map{3} & 97.0\% (\(0.85\))          & 60.4\% (\(0.69\))          & 45.7\% (\(0.66\))          & 32.1\% (\(0.63\))          & 2.5\% (\(0.57\))          & 0.1\%              \\
        \Map{4} & 94.7\% (\(0.61\))          & 68.2\% (\(0.39\))          & 46.9\% (\(0.36\))          & 35.6\% (\(0.35\))          & 8.7\% (\(0.32\))          & 9.2\%              \\
        \Map{5} & 96.8\% (\(0.53\))          & 81.7\% (\(0.21\))          & 50.1\% (\(0.15\))          & 36.7\% (\(0.11\))          & 4.9\% (\(0.04\))          & 9.1\%              \\
        \Map{6} & 93.5\% (\(0.43\))          & 61.7\% (\(0.21\))          & 43.5\% (\(0.18\))          & 21.4\% (\(0.14\))          & 2.0\% (\(0.05\))          & 7.3\%              \\
        \Map{7} & 92.5\% (\(0.33\))          & 60.2\% (\(0.13\))          & 52.1\% (\(0.11\))          & 15.4\% (\(0.07\))          & 2.7\% (\(0.04\))          & 7.8\%              \\
        \Map{8} & 94.6\% (\(0.35\))          & 71.0\% (\(0.11\))          & 37.0\% (\(0.06\))          & 15.3\% (\(0.04\))          & 3.0\% (\(0.02\))          & 7.6\%              \\
        \bottomrule
    \end{tabular*}
\end{table}

All algorithms are implemented in C++20 and compiled with GCC 15.2 at \texttt{-O3 -march=native} on five identical \texttt{x86\_64} Linux instances with 4~GB of RAM each.
To eliminate timer overhead and clock jitter at sub-microsecond scales, queries are timed in 128-query chunks via \texttt{std::chrono::steady\_clock} over 8,192 query pairs following 128 unmeasured warm-up queries.
Unless otherwise stated, \texttt{H-BRICK} is configured with \texttt{HBrickSkipLift} at \((b=24, g=4)\).

\subsection{Hierarchy and Parameter Trade-offs}
\label{subsec:hierarchy-configurations}

The construction exposes two numerical knobs: the base-tile side \(b\) and the aggregation factor \(g\).
Table~\ref{tab:param-map} reports \texttt{HBrickSkipLift} preprocessing time, query latency, and index memory for the complete factorial sweep \(b \in \{4, 8, 16, 24, 32, 48, 64, 96\}\) \(\times\) \(g \in \{2, 4, 8\}\) on every benchmark map, and Table~\ref{tab:param-regime} presents the same measurements aggregated by orientation regime.
Map entries are medians over the six regimes of the per-regime measurement, regime entries are medians over the eight maps, and query entries are built from the per-regime mean latencies.

Throughout this subsection, every quoted value is a median over the individual runs---six orientation regimes for map entries and eight benchmark maps for regime entries---and the stated ranges are the extremes of those medians rather than readings from single runs; query entries are built from the per-run mean latencies.

Query latency is the least parameter-sensitive of the three quantities.
Across the entire sweep it remains within \(0.20\)--\(4.67\,\mu\mathrm{s}\), and on each map it varies by less than a factor of five: \Map{1} spans \(0.20\)--\(0.98\,\mu\mathrm{s}\) and the largest map \Map{8} only \(2.91\)--\(4.67\,\mu\mathrm{s}\) over all \(b\) and \(g\).
Preprocessing and memory are far more sensitive.
Fine tiles (\(b=4\)) multiply the number of boundary ports lifted at each level: the median index footprint reaches \(901.5\,\mathrm{MiB}\) on \Map{8}, \(111.0\,\mathrm{MiB}\) on \Map{3}, and \(77.7\,\mathrm{MiB}\) on \Map{1}, without improving query latency.
Large tiles (\(b \ge 48\)) let the intra-tile closure term \(O(Kb^4\log b)\)~\eqref{eq:grid-preprocessing} dominate: on \Map{8} at \(g=4\), preprocessing grows from \(2.40\,\mathrm{s}\) at \(b=16\) to \(11.96\,\mathrm{s}\) at \(b=64\) and \(30.91\,\mathrm{s}\) at \(b=96\), while the median footprint rises from \(315.5\)--\(333.6\,\mathrm{MiB}\) over \(b \in \{16,24,32\}\) to \(840.3\,\mathrm{MiB}\) at \(b=96\).
The memory footprint therefore has a broad minimum at \(b \in \{16, 24, 32\}\), where \Map{8} needs about a third of a gibibyte and \Map{1} and \Map{3} about \(28\) and \(31\,\mathrm{MiB}\).
Wide aggregation (\(g=8\)) inflates interface width, pushing \Map{8} preprocessing to \(6.04\)--\(30.56\,\mathrm{s}\) across the sweep, whereas \(g=2\) deepens the hierarchy without lowering query latency.
Within this region, Table~\ref{tab:param-regime} places \((b=24, g=4)\) among the three smallest preprocessing times and the three smallest index footprints in every orientation regime, and Table~\ref{tab:param-map} places it among the three smallest footprints on six of the eight maps.
We adopt it as the operating point, and, unless otherwise stated, every later comparison with the other methods uses this configuration.
The same cell is not a query-latency minimum.
At \(g=4\), raising the base tile from \(b=24\) to \(b=64\) multiplies the regime-median preprocessing time by a factor of about \(11\)--\(14\), while the regime-median query latency stays within a factor of \(1.5\), so the added construction cost is not repaid by the query.
The chosen point lies on that plateau: between \(b=16\) and \(b=24\) the regime-median index footprint changes by about one mebibyte.
At \((b=24, g=4)\) the per-map medians span \(0.06\)--\(3.52\,\mathrm{s}\) preprocessing, \(7.0\)--\(315.5\,\mathrm{MiB}\) index memory, and \(0.28\)--\(3.03\,\mu\mathrm{s}\) query latency.

Aggregated over the eight maps, the regime medians confirm the orientation-insensitivity of the hierarchy: for a fixed \((b,g)\) the median query latency moves by well under a factor of two between \(V01\) and \(V06\), and at \((b=24, g=4)\) it stays between \(0.85\) and \(1.06\,\mu\mathrm{s}\) across the six regimes (Table~\ref{tab:param-regime}), with a median index footprint between \(48.5\) and \(50.1\,\mathrm{MiB}\) and median preprocessing between \(0.15\) and \(0.28\,\mathrm{s}\).

The latency of an individual query is governed by the level of the lowest ancestor that certifies it, not by graph diameter (Section~\ref{sec:analysis}).
A pair whose endpoints share a base tile is certified by a single local closure lookup \((A_T^\star)_{st}\) when that entry is one; a zero entry is not conclusive, because the pair may still be connected through a path that leaves the tile, so the query proceeds up the ancestor chain exactly as for pairs in different tiles.

Consequently, unreachable pairs are always evaluated up to the root and form the slow tail of every latency distribution reported here, and the residual max-to-min variation of \texttt{H-BRICK} in Figure~\ref{fig:query_vs_orientation} is consistent with the changing share of unreachable pairs across regimes rather than with any change in map diameter.

\begin{table}[H]
    \caption{Preprocessing time, query latency, and index memory of \texttt{HBrickSkipLift} for every benchmark map across base-tile side \(b \in \{4, 8, 16, 24, 32, 48, 64, 96\}\) and aggregation factor \(g \in \{2, 4, 8\}\). Each entry is the median across the six orientation regimes of the per-regime measurement; query entries are the per-regime mean query latencies. Bold entries mark the three smallest values of the corresponding metric within each map block, and the smallest is also underlined.}
    \label{tab:param-map}
    \centering
    \scriptsize
    \setlength{\tabcolsep}{1.3pt}
    \renewcommand{\arraystretch}{1.02}
    \newcolumntype{Z}{w{r}{2.7em}}
    \begin{adjustwidth}{-\extralength}{0cm}
    \resizebox{\fulllength}{!}{%
    \begin{tabular}{c c ZZZZZZZZ|ZZZZZZZZ|ZZZZZZZZ}
        \toprule
         &  & \multicolumn{8}{c}{Preprocessing (s)} & \multicolumn{8}{c}{Query latency (\(\mu\mathrm{s}\))} & \multicolumn{8}{c}{Index memory (MiB)} \\
        \cmidrule(lr){3-10} \cmidrule(lr){11-18} \cmidrule(lr){19-26}
        Map & \(g\) & \(4\) & \(8\) & \(16\) & \(24\) & \(32\) & \(48\) & \(64\) & \(96\) & \(4\) & \(8\) & \(16\) & \(24\) & \(32\) & \(48\) & \(64\) & \(96\) & \(4\) & \(8\) & \(16\) & \(24\) & \(32\) & \(48\) & \(64\) & \(96\) \\
        \midrule
        \multirow{3}{*}{\Map{1}} & \(2\) & 0.26 & 0.13 & 0.08 & \textbf{0.07} & 0.17 & 0.37 & 0.45 & 0.83 & 0.98 & 0.62 & 0.32 & 0.28 & 0.23 & \textbf{0.22} & \textbf{0.21} & 0.22 & 104.2 & 50.3 & 31.2 & 30.0 & 29.2 & 37.4 & 50.3 & 85.1 \\
         & \(4\) & 0.20 & 0.11 & \textbf{0.07} & \underline{\textbf{0.06}} & 0.18 & 0.39 & 0.47 & 0.83 & 0.87 & 0.44 & 0.30 & 0.28 & 0.24 & 0.22 & \underline{\textbf{0.20}} & 0.24 & 77.7 & 38.2 & \textbf{27.8} & \underline{\textbf{27.7}} & \textbf{27.9} & 36.2 & 50.1 & 85.0 \\
         & \(8\) & 0.19 & 0.12 & 0.09 & 0.07 & 0.17 & 0.39 & 0.46 & 0.80 & 0.83 & 0.45 & 0.29 & 0.25 & 0.22 & 0.24 & 0.24 & 0.24 & 74.2 & 39.3 & 28.2 & 28.1 & 29.8 & 37.7 & 51.8 & 85.6 \\
        \midrule
        \multirow{3}{*}{\Map{2}} & \(2\) & 1.39 & \textbf{0.79} & \underline{\textbf{0.47}} & 1.07 & 1.04 & 2.29 & 4.56 & 10.27 & 2.20 & 2.20 & 2.03 & 2.36 & 1.90 & 2.44 & 2.00 & 2.39 & 306.4 & 170.5 & 107.6 & 106.3 & \underline{\textbf{98.1}} & 130.4 & 174.3 & 306.8 \\
         & \(4\) & 1.38 & 0.85 & \textbf{0.67} & 1.22 & 1.30 & 2.44 & 4.57 & 10.18 & 2.15 & 2.70 & 1.90 & 1.89 & 2.23 & \textbf{1.74} & \textbf{1.82} & 2.31 & 236.9 & 134.0 & \textbf{100.3} & 101.3 & \textbf{98.7} & 127.8 & 177.6 & 308.6 \\
         & \(8\) & 2.53 & 2.41 & 1.43 & 1.71 & 2.06 & 2.89 & 5.41 & 11.11 & 2.27 & 3.62 & 1.98 & \underline{\textbf{1.62}} & 1.87 & 2.89 & 3.12 & 2.13 & 263.6 & 175.1 & 123.8 & 119.3 & 130.0 & 150.6 & 208.2 & 320.2 \\
        \midrule
        \multirow{3}{*}{\Map{3}} & \(2\) & 0.44 & 0.10 & \textbf{0.08} & 0.09 & 0.72 & 1.33 & 2.83 & 7.22 & 1.38 & 0.73 & 0.51 & 0.51 & 0.44 & 0.46 & 0.39 & 0.39 & 145.4 & 37.5 & 32.7 & 36.5 & 43.4 & 65.8 & 101.8 & 186.4 \\
         & \(4\) & 0.37 & 0.09 & \underline{\textbf{0.08}} & 0.09 & 0.73 & 1.34 & 2.78 & 7.73 & 1.26 & 0.64 & 0.47 & 0.45 & 0.42 & 0.38 & \textbf{0.37} & 0.40 & 111.0 & 31.6 & \textbf{30.9} & 35.3 & 42.6 & 65.2 & 101.7 & 186.2 \\
         & \(8\) & 0.41 & 0.09 & \textbf{0.08} & 0.09 & 0.75 & 1.35 & 2.77 & 7.34 & 1.22 & 0.51 & 0.44 & 0.39 & 0.39 & 0.42 & \textbf{0.36} & \underline{\textbf{0.32}} & 109.1 & \textbf{30.7} & \underline{\textbf{30.5}} & 35.0 & 42.9 & 65.4 & 101.8 & 186.2 \\
        \midrule
        \multirow{3}{*}{\Map{4}} & \(2\) & 0.88 & 0.47 & \textbf{0.28} & \underline{\textbf{0.25}} & 0.86 & 1.46 & 2.78 & 7.31 & 1.85 & 1.63 & 1.35 & 1.42 & 1.20 & 1.30 & 1.14 & 1.37 & 220.5 & 114.7 & 71.2 & 66.4 & 65.2 & 84.8 & 118.3 & 206.3 \\
         & \(4\) & 0.88 & 0.49 & 0.40 & \textbf{0.32} & 1.06 & 1.47 & 2.80 & 7.37 & 1.87 & 1.63 & 1.41 & 1.19 & 1.09 & 1.04 & \textbf{1.02} & 1.32 & 169.0 & 88.3 & \textbf{64.7} & \underline{\textbf{62.3}} & \textbf{63.8} & 82.6 & 119.3 & 206.8 \\
         & \(8\) & 1.49 & 1.45 & 0.71 & 0.58 & 1.31 & 1.79 & 3.40 & 7.46 & 1.88 & 1.78 & 1.16 & \underline{\textbf{1.00}} & 1.03 & 1.17 & 1.10 & \textbf{1.01} & 177.5 & 104.3 & 73.5 & 69.3 & 77.0 & 92.3 & 131.8 & 211.7 \\
        \midrule
        \multirow{3}{*}{\Map{5}} & \(2\) & 0.09 & 0.05 & \underline{\textbf{0.03}} & 0.05 & 0.08 & 0.15 & 0.26 & 0.56 & 0.96 & 0.84 & 0.59 & 0.62 & 0.59 & 0.56 & 0.54 & 0.42 & 25.6 & 13.6 & 8.5 & 7.6 & 7.7 & 9.0 & 11.3 & 17.8 \\
         & \(4\) & 0.07 & 0.05 & \textbf{0.03} & 0.06 & 0.09 & 0.15 & 0.26 & 0.59 & 0.73 & 0.58 & 0.52 & 0.52 & 0.45 & 0.41 & 0.39 & \textbf{0.32} & 18.6 & 10.2 & \textbf{7.3} & \underline{\textbf{7.0}} & \textbf{7.1} & 8.8 & 11.1 & 17.8 \\
         & \(8\) & 0.12 & 0.08 & \textbf{0.04} & 0.08 & 0.10 & 0.16 & 0.25 & 0.60 & 0.79 & 0.54 & 0.46 & 0.47 & 0.45 & 0.35 & \textbf{0.32} & \underline{\textbf{0.30}} & 18.1 & 10.2 & 7.4 & 7.7 & 7.6 & 8.9 & 11.0 & 17.8 \\
        \midrule
        \multirow{3}{*}{\Map{6}} & \(2\) & 0.18 & 0.10 & \underline{\textbf{0.06}} & \textbf{0.07} & 0.18 & 0.29 & 0.62 & 1.19 & 1.35 & 0.93 & 0.69 & 0.85 & 0.66 & 0.70 & 0.57 & 0.62 & 45.0 & 24.3 & 15.1 & 14.5 & 14.0 & 17.8 & 24.9 & 39.5 \\
         & \(4\) & 0.17 & 0.10 & \textbf{0.09} & 0.09 & 0.20 & 0.30 & 0.66 & 1.16 & 1.05 & 0.88 & 0.64 & 0.60 & 0.56 & 0.63 & 0.44 & \underline{\textbf{0.38}} & 32.3 & 20.4 & \textbf{13.8} & \underline{\textbf{13.2}} & \textbf{13.9} & 17.7 & 25.2 & 39.4 \\
         & \(8\) & 0.31 & 0.19 & 0.15 & 0.13 & 0.31 & 0.35 & 0.64 & 1.15 & 1.05 & 0.74 & 0.60 & 0.73 & 0.56 & 0.45 & \textbf{0.42} & \textbf{0.39} & 34.6 & 21.6 & 17.1 & 15.8 & 17.4 & 18.9 & 25.2 & 39.4 \\
        \midrule
        \multirow{3}{*}{\Map{7}} & \(2\) & 0.96 & 0.56 & \underline{\textbf{0.31}} & \textbf{0.53} & 0.79 & 1.41 & 2.62 & 6.14 & 1.98 & 1.88 & 1.56 & 1.59 & 1.36 & 1.57 & 1.31 & 1.42 & 236.4 & 126.8 & 77.2 & 73.3 & \textbf{67.3} & 84.8 & 112.1 & 188.0 \\
         & \(4\) & 0.93 & 0.55 & \textbf{0.45} & 0.61 & 0.90 & 1.52 & 2.85 & 6.43 & 2.01 & 1.62 & 1.33 & 1.47 & \underline{\textbf{1.15}} & 1.30 & 1.30 & 1.45 & 182.0 & 98.2 & 70.4 & \textbf{68.6} & \underline{\textbf{66.5}} & 82.6 & 113.6 & 188.7 \\
         & \(8\) & 1.58 & 1.42 & 0.84 & 0.91 & 1.35 & 1.82 & 3.27 & 6.56 & 1.74 & 2.12 & 1.36 & 1.26 & \textbf{1.19} & 1.42 & 1.41 & \textbf{1.16} & 192.6 & 119.6 & 82.8 & 78.5 & 83.2 & 95.4 & 130.4 & 195.1 \\
        \midrule
        \multirow{3}{*}{\Map{8}} & \(2\) & 5.73 & \textbf{2.96} & \underline{\textbf{1.71}} & 3.02 & 3.32 & 6.86 & 11.37 & 29.13 & 3.00 & 3.05 & 3.25 & 3.37 & 3.14 & 3.65 & 3.69 & 4.19 & 1314.6 & 676.2 & 397.6 & 371.5 & \textbf{326.4} & 402.5 & 499.3 & 842.4 \\
         & \(4\) & 4.94 & 3.04 & \textbf{2.40} & 3.52 & 3.84 & 6.88 & 11.96 & 30.91 & 3.42 & \underline{\textbf{2.91}} & 3.05 & 3.03 & \textbf{2.98} & 3.61 & 3.58 & 3.88 & 901.5 & 547.0 & 333.6 & \underline{\textbf{315.5}} & \textbf{319.8} & 397.2 & 505.9 & 840.3 \\
         & \(8\) & 7.65 & 7.17 & 10.98 & 7.07 & 6.04 & 8.94 & 15.98 & 30.56 & 3.14 & \textbf{2.91} & 4.67 & 3.70 & 3.42 & 3.33 & 3.34 & 4.11 & 954.8 & 640.9 & 450.4 & 396.3 & 386.1 & 448.3 & 580.2 & 895.9 \\
        \bottomrule
    \end{tabular}}
    \end{adjustwidth}
\end{table}

\begin{table}[H]
    \caption{Preprocessing time, query latency, and index memory of \texttt{HBrickSkipLift} for each orientation regime across base-tile side \(b \in \{4, 8, 16, 24, 32, 48, 64, 96\}\) and aggregation factor \(g \in \{2, 4, 8\}\). Each entry is the median across the eight benchmark maps of the per-map measurement. Bold entries mark the three smallest values of the corresponding metric within each regime block, and the smallest is also underlined.}
    \label{tab:param-regime}
    \centering
    \scriptsize
    \setlength{\tabcolsep}{1.3pt}
    \renewcommand{\arraystretch}{1.02}
    \newcolumntype{Y}{w{r}{2.7em}}
    \begin{adjustwidth}{-\extralength}{0cm}
    \resizebox{\fulllength}{!}{%
    \begin{tabular}{c c YYYYYYYY|YYYYYYYY|YYYYYYYY}
        \toprule
         &  & \multicolumn{8}{c}{Preprocessing (s)} & \multicolumn{8}{c}{Query latency (\(\mu\mathrm{s}\))} & \multicolumn{8}{c}{Index memory (MiB)} \\
        \cmidrule(lr){3-10} \cmidrule(lr){11-18} \cmidrule(lr){19-26}
        Map & \(g\) & \(4\) & \(8\) & \(16\) & \(24\) & \(32\) & \(48\) & \(64\) & \(96\) & \(4\) & \(8\) & \(16\) & \(24\) & \(32\) & \(48\) & \(64\) & \(96\) & \(4\) & \(8\) & \(16\) & \(24\) & \(32\) & \(48\) & \(64\) & \(96\) \\
        \midrule
        \multirow{3}{*}{\(V01\)} & \(2\) & 1.18 & 0.50 & \textbf{0.25} & \underline{\textbf{0.24}} & 0.86 & 1.65 & 3.49 & 8.95 & 1.59 & 1.43 & 1.26 & 1.15 & 1.00 & 1.07 & 0.97 & 1.15 & 184.0 & 83.6 & 53.1 & \textbf{52.8} & 55.7 & 76.8 & 108.4 & 188.7 \\
         & \(4\) & 0.98 & 0.45 & 0.35 & \textbf{0.28} & 0.95 & 1.62 & 3.69 & 9.18 & 1.44 & 1.16 & 1.20 & 1.06 & 0.96 & 0.95 & 1.05 & 1.12 & 141.0 & 64.3 & \underline{\textbf{48.9}} & \textbf{50.1} & 54.6 & 75.4 & 109.1 & 189.0 \\
         & \(8\) & 1.53 & 1.17 & 0.60 & 0.50 & 1.27 & 1.91 & 4.33 & 9.26 & 1.46 & 1.01 & 0.92 & 0.92 & \textbf{0.86} & 1.02 & \underline{\textbf{0.70}} & \textbf{0.79} & 144.4 & 72.9 & 53.1 & 53.5 & 61.3 & 80.4 & 117.6 & 192.2 \\
        \midrule
        \multirow{3}{*}{\(V02\)} & \(2\) & 0.80 & 0.34 & \textbf{0.22} & \underline{\textbf{0.19}} & 0.78 & 1.42 & 2.94 & 7.50 & 1.70 & 1.28 & 1.09 & 1.07 & 0.86 & 0.94 & 0.85 & 0.95 & 183.1 & 82.6 & 52.1 & \textbf{51.6} & 54.5 & 75.5 & 107.2 & 187.4 \\
         & \(4\) & 0.74 & 0.35 & 0.28 & \textbf{0.24} & 0.88 & 1.46 & 3.27 & 7.66 & 1.66 & 1.11 & 1.07 & 0.85 & 0.79 & 0.81 & 0.85 & 0.75 & 140.1 & 63.4 & \underline{\textbf{47.9}} & \textbf{48.9} & 53.4 & 74.1 & 107.8 & 187.7 \\
         & \(8\) & 1.17 & 0.95 & 0.53 & 0.43 & 1.17 & 1.68 & 3.43 & 7.72 & 1.51 & 1.03 & 0.91 & 0.81 & \textbf{0.72} & \textbf{0.69} & 0.74 & \underline{\textbf{0.62}} & 143.4 & 71.9 & 52.2 & 52.4 & 60.1 & 79.1 & 116.3 & 190.9 \\
        \midrule
        \multirow{3}{*}{\(V03\)} & \(2\) & 0.70 & 0.32 & \textbf{0.19} & \underline{\textbf{0.17}} & 0.78 & 1.42 & 2.75 & 7.07 & 1.66 & 1.35 & 1.02 & 1.16 & 0.93 & 1.05 & 0.91 & 1.05 & 183.0 & 82.5 & 52.0 & \textbf{51.5} & 54.3 & 75.3 & 107.1 & 187.3 \\
         & \(4\) & 0.65 & 0.33 & 0.25 & \textbf{0.21} & 0.85 & 1.43 & 2.88 & 7.52 & 1.65 & 1.30 & 0.98 & 1.02 & 0.84 & 0.95 & \textbf{0.74} & 0.94 & 140.0 & 63.3 & \underline{\textbf{47.8}} & \textbf{48.8} & 53.2 & 73.9 & 107.7 & 187.6 \\
         & \(8\) & 0.99 & 0.81 & 0.47 & 0.36 & 1.05 & 1.63 & 3.08 & 7.49 & 1.51 & 1.32 & 0.91 & 0.88 & \textbf{0.77} & 0.92 & 0.84 & \underline{\textbf{0.68}} & 143.3 & 71.8 & 52.0 & 52.2 & 60.0 & 78.9 & 116.2 & 190.8 \\
        \midrule
        \multirow{3}{*}{\(V04\)} & \(2\) & 0.63 & 0.28 & \textbf{0.18} & \underline{\textbf{0.17}} & 0.75 & 1.32 & 2.59 & 6.07 & 1.67 & 1.23 & 1.01 & 1.15 & 0.89 & 1.05 & 0.84 & 0.94 & 183.0 & 82.5 & 52.0 & \textbf{51.4} & 54.3 & 75.0 & 106.8 & 187.0 \\
         & \(4\) & 0.58 & 0.28 & 0.24 & \textbf{0.21} & 0.78 & 1.39 & 2.65 & 6.35 & 1.87 & 1.23 & 1.04 & 1.03 & 0.82 & 0.81 & \underline{\textbf{0.69}} & 0.96 & 140.0 & 63.3 & \underline{\textbf{47.8}} & \textbf{48.8} & 53.2 & 73.7 & 107.4 & 187.2 \\
         & \(8\) & 0.89 & 0.68 & 0.39 & 0.35 & 1.01 & 1.46 & 2.84 & 6.36 & 1.57 & 1.21 & 0.89 & 0.87 & 0.89 & 0.79 & \textbf{0.76} & \textbf{0.72} & 143.3 & 71.8 & 52.0 & 52.2 & 59.9 & 78.9 & 115.9 & 190.4 \\
        \midrule
        \multirow{3}{*}{\(V05\)} & \(2\) & 0.55 & 0.26 & \textbf{0.16} & \underline{\textbf{0.15}} & 0.70 & 1.19 & 2.19 & 5.19 & 1.53 & 1.30 & 1.06 & 0.98 & 0.87 & 0.87 & 0.70 & 0.71 & 182.9 & 82.4 & 51.9 & \textbf{51.3} & 54.1 & 74.8 & 106.6 & 186.8 \\
         & \(4\) & 0.48 & 0.25 & 0.21 & \textbf{0.17} & 0.73 & 1.21 & 2.33 & 5.24 & 1.34 & 1.20 & 0.85 & 0.88 & 0.71 & 0.76 & \textbf{0.69} & \textbf{0.66} & 139.9 & 63.2 & \underline{\textbf{47.7}} & \textbf{48.6} & 53.1 & 73.5 & 107.2 & 187.1 \\
         & \(8\) & 0.73 & 0.50 & 0.30 & 0.26 & 0.88 & 1.30 & 2.37 & 4.97 & 1.51 & 1.42 & 0.86 & 0.88 & 0.82 & 0.77 & 0.89 & \underline{\textbf{0.65}} & 143.2 & 71.7 & 51.9 & 52.1 & 59.8 & 78.8 & 115.7 & 190.3 \\
        \midrule
        \multirow{3}{*}{\(V06\)} & \(2\) & 0.53 & 0.24 & \textbf{0.15} & \underline{\textbf{0.14}} & 0.59 & 0.96 & 1.61 & 3.62 & 1.41 & 1.34 & 0.97 & 1.19 & 1.15 & 0.96 & 0.92 & 1.02 & 181.5 & 82.0 & 51.7 & \textbf{51.1} & 54.1 & 75.1 & 106.9 & 187.0 \\
         & \(4\) & 0.44 & 0.23 & 0.18 & \textbf{0.15} & 0.61 & 0.97 & 1.61 & 3.77 & 1.45 & 1.33 & 0.95 & 0.92 & 0.92 & 0.82 & \underline{\textbf{0.75}} & \textbf{0.77} & 138.6 & 62.8 & \underline{\textbf{47.5}} & \textbf{48.5} & 53.0 & 73.7 & 107.5 & 187.3 \\
         & \(8\) & 0.66 & 0.47 & 0.27 & 0.22 & 0.71 & 1.05 & 1.73 & 3.79 & 1.36 & 1.39 & 1.00 & 0.94 & \textbf{0.78} & 0.87 & 0.80 & 0.82 & 141.9 & 71.3 & 51.7 & 51.9 & 59.7 & 78.7 & 116.0 & 190.5 \\
        \bottomrule
    \end{tabular}}
    \end{adjustwidth}
\end{table}

\subsection{Hierarchy Ablation: Flat vs.\ Multi-Level Composition}
\label{subsec:hierarchy-ablation}
\renewcommand{\topfraction}{0.85}
\renewcommand{\bottomfraction}{0.7}
\renewcommand{\textfraction}{0.07}
\renewcommand{\floatpagefraction}{0.5}
\begin{table}[t]
    \caption{Preprocessing time, query latency, and index memory of single-level Flat BRICK (\texttt{BrickSearch}) at \(g=4\) for every benchmark map across base-tile side \(b\). The upper block lists the Flat BRICK measurements; the lower block lists the Flat-to-H-BRICK ratios (\(\times\), values above one favour the hierarchy). Both are averaged over the six orientation regimes. Bold entries mark the three best values of each quantity within a map block, and the best is also underlined.}
    \label{tab:flat-map}
    \centering
    \scriptsize
    \setlength{\tabcolsep}{1.0pt}
    \renewcommand{\arraystretch}{1.02}
    \newcolumntype{X}{w{r}{2.9em}}
    \begin{adjustwidth}{-\extralength}{0cm}
    \resizebox{\fulllength}{!}{%
    \begin{tabular}{c XXXXXXXX|XXXXXXXX|XXXXXXXX}
        \toprule
         & \multicolumn{8}{c}{Preprocessing (s)} & \multicolumn{8}{c}{Query latency (\(\mu\mathrm{s}\))} & \multicolumn{8}{c}{Index memory (MiB)} \\
        \cmidrule(lr){2-9} \cmidrule(lr){10-17} \cmidrule(lr){18-25}
        Map & \(4\) & \(8\) & \(16\) & \(24\) & \(32\) & \(48\) & \(64\) & \(96\) & \(4\) & \(8\) & \(16\) & \(24\) & \(32\) & \(48\) & \(64\) & \(96\) & \(4\) & \(8\) & \(16\) & \(24\) & \(32\) & \(48\) & \(64\) & \(96\) \\
        \midrule
        \Map{1} & 0.03 & \textbf{0.03} & \underline{\textbf{0.03}} & \textbf{0.03} & 0.15 & 0.39 & 0.52 & 1.04 & 67.92 & 31.83 & 13.04 & 8.52 & 5.42 & \textbf{3.09} & \textbf{2.00} & \underline{\textbf{1.58}} & 26.6 & \textbf{19.8} & \underline{\textbf{18.8}} & \textbf{21.3} & 24.0 & 33.5 & 48.2 & 83.5 \\
        \Map{2} & \underline{\textbf{0.08}} & \textbf{0.09} & \textbf{0.09} & 0.72 & 0.82 & 1.99 & 4.01 & 9.45 & 1110 & 577 & 322 & 249 & 213 & \textbf{145} & \textbf{117} & \underline{\textbf{80.22}} & \textbf{55.5} & \underline{\textbf{43.8}} & \textbf{44.1} & 56.2 & 67.1 & 105.6 & 160.1 & 295.1 \\
        \Map{3} & \textbf{0.05} & \underline{\textbf{0.04}} & \textbf{0.05} & 0.06 & 0.68 & 1.24 & 2.63 & 6.91 & 528 & 45.11 & 19.91 & 10.39 & 7.01 & \textbf{4.09} & \textbf{3.07} & \underline{\textbf{2.43}} & 40.8 & \underline{\textbf{24.4}} & \textbf{27.7} & \textbf{33.1} & 41.4 & 64.4 & 101.1 & 185.8 \\
        \Map{4} & 0.08 & \underline{\textbf{0.07}} & \textbf{0.07} & \textbf{0.07} & 0.75 & 1.28 & 2.78 & 7.21 & 872 & 441 & 235 & 152 & 127 & \textbf{79.71} & \textbf{63.79} & \underline{\textbf{52.57}} & 43.7 & \underline{\textbf{33.6}} & \textbf{34.0} & \textbf{38.8} & 48.2 & 72.2 & 111.2 & 200.8 \\
        \Map{5} & \underline{\textbf{0.01}} & \textbf{0.01} & \textbf{0.01} & 0.04 & 0.07 & 0.13 & 0.22 & 0.54 & 77.92 & 37.89 & 19.01 & 16.17 & 12.01 & \textbf{9.37} & \textbf{8.14} & \underline{\textbf{5.47}} & 6.2 & \textbf{4.6} & \underline{\textbf{4.4}} & \textbf{5.2} & 5.8 & 8.0 & 10.4 & 17.5 \\
        \Map{6} & \textbf{0.02} & \underline{\textbf{0.02}} & \textbf{0.02} & 0.04 & 0.16 & 0.27 & 0.60 & 1.15 & 160 & 81.74 & 39.88 & 28.22 & 21.82 & \textbf{18.11} & \textbf{10.45} & \underline{\textbf{7.78}} & 10.7 & \textbf{8.3} & \underline{\textbf{8.0}} & \textbf{9.2} & 11.0 & 15.5 & 23.9 & 38.7 \\
        \Map{7} & \underline{\textbf{0.07}} & \textbf{0.07} & \textbf{0.07} & 0.31 & 0.62 & 1.24 & 2.60 & 6.12 & 903 & 442 & 284 & 187 & 135 & \textbf{107} & \textbf{82.53} & \underline{\textbf{66.46}} & 44.2 & \textbf{34.5} & \underline{\textbf{33.9}} & \textbf{40.3} & 47.5 & 69.3 & 103.6 & 181.2 \\
        \Map{8} & \underline{\textbf{0.27}} & \textbf{0.29} & \textbf{0.29} & 1.82 & 2.44 & 5.71 & 11.13 & 29.16 & 5784 & 3044 & 1592 & 1186 & 969 & \textbf{711} & \textbf{622} & \underline{\textbf{436}} & 179.5 & \textbf{140.1} & \underline{\textbf{139.2}} & \textbf{170.7} & 200.6 & 301.6 & 438.6 & 793.3 \\
        \midrule
        \textbf{Ratio} (\(\times\)) & \multicolumn{8}{c|}{Preprocessing} & \multicolumn{8}{c|}{Query speedup} & \multicolumn{8}{c}{Index memory} \\
        \midrule
        \Map{1} & 0.16 & 0.25 & 0.38 & 0.43 & 0.81 & \textbf{0.90} & \textbf{0.93} & \underline{\textbf{0.94}} & \underline{\textbf{72.8}} & \textbf{56.6} & \textbf{33.1} & 21.5 & 17.0 & 10.1 & 6.57 & 5.00 & 0.34 & 0.52 & 0.68 & 0.77 & 0.86 & \textbf{0.93} & \textbf{0.96} & \underline{\textbf{0.98}} \\
        \Map{2} & 0.06 & 0.10 & 0.14 & 0.61 & 0.64 & \textbf{0.86} & \textbf{0.92} & \underline{\textbf{0.97}} & \underline{\textbf{506}} & \textbf{227} & \textbf{163} & 118 & 97.5 & 80.3 & 63.1 & 35.8 & 0.23 & 0.33 & 0.44 & 0.55 & 0.68 & \textbf{0.82} & \textbf{0.90} & \underline{\textbf{0.96}} \\
        \Map{3} & 0.15 & 0.49 & 0.62 & 0.67 & 0.97 & \underline{\textbf{0.98}} & \textbf{0.98} & \textbf{0.98} & \underline{\textbf{391}} & \textbf{71.1} & \textbf{42.4} & 24.3 & 17.3 & 11.0 & 8.38 & 6.29 & 0.37 & 0.77 & 0.90 & 0.94 & 0.97 & \textbf{0.99} & \textbf{0.99} & \underline{\textbf{1.00}} \\
        \Map{4} & 0.09 & 0.14 & 0.18 & 0.23 & 0.75 & \textbf{0.88} & \textbf{0.94} & \underline{\textbf{0.97}} & \underline{\textbf{478}} & \textbf{280} & \textbf{164} & 128 & 107 & 75.6 & 58.7 & 39.4 & 0.26 & 0.38 & 0.52 & 0.62 & 0.75 & \textbf{0.87} & \textbf{0.93} & \underline{\textbf{0.97}} \\
        \Map{5} & 0.11 & 0.19 & 0.26 & 0.64 & 0.80 & \textbf{0.91} & \textbf{0.92} & \underline{\textbf{0.97}} & \underline{\textbf{102}} & \textbf{62.1} & \textbf{35.2} & 31.0 & 25.3 & 22.2 & 20.2 & 17.9 & 0.33 & 0.45 & 0.60 & 0.74 & 0.81 & \textbf{0.90} & \textbf{0.94} & \underline{\textbf{0.98}} \\
        \Map{6} & 0.10 & 0.15 & 0.20 & 0.43 & 0.80 & \textbf{0.91} & \textbf{0.93} & \underline{\textbf{0.97}} & \underline{\textbf{153}} & \textbf{91.1} & \textbf{62.0} & 41.8 & 39.3 & 27.8 & 24.0 & 20.5 & 0.33 & 0.41 & 0.58 & 0.69 & 0.79 & \textbf{0.87} & \textbf{0.95} & \underline{\textbf{0.98}} \\
        \Map{7} & 0.07 & 0.12 & 0.16 & 0.51 & 0.70 & \textbf{0.85} & \textbf{0.92} & \underline{\textbf{0.97}} & \underline{\textbf{443}} & \textbf{259} & \textbf{202} & 127 & 115 & 76.9 & 59.3 & 47.0 & 0.24 & 0.35 & 0.48 & 0.58 & 0.71 & \textbf{0.83} & \textbf{0.91} & \underline{\textbf{0.96}} \\
        \Map{8} & 0.05 & 0.08 & 0.11 & 0.51 & 0.60 & \textbf{0.81} & \textbf{0.88} & \underline{\textbf{0.96}} & \underline{\textbf{1657}} & \textbf{1051} & \textbf{515} & 382 & 333 & 204 & 167 & 116 & 0.20 & 0.25 & 0.41 & 0.54 & 0.62 & \textbf{0.75} & \textbf{0.86} & \underline{\textbf{0.94}} \\
        \bottomrule
    \end{tabular}}
    \end{adjustwidth}
\end{table}

\begin{table}[t]
    \caption{Preprocessing time, query latency, and index memory of single-level Flat BRICK (\texttt{BrickSearch}) at \(g=4\) for each orientation regime across base-tile side \(b\). The upper block lists the Flat BRICK measurements; the lower block lists the Flat-to-H-BRICK ratios (\(\times\), values above one favour the hierarchy). Both are averaged over the eight benchmark maps. Bold entries mark the three best values of each quantity within a regime block, and the best is also underlined.}
    \label{tab:flat-regime}
    \centering
    \scriptsize
    \setlength{\tabcolsep}{1.0pt}
    \renewcommand{\arraystretch}{1.02}
    \newcolumntype{W}{w{r}{2.9em}}
    \begin{adjustwidth}{-\extralength}{0cm}
    \resizebox{\fulllength}{!}{%
    \begin{tabular}{c WWWWWWWW|WWWWWWWW|WWWWWWWW}
        \toprule
         & \multicolumn{8}{c}{Preprocessing (s)} & \multicolumn{8}{c}{Query latency (\(\mu\mathrm{s}\))} & \multicolumn{8}{c}{Index memory (MiB)} \\
        \cmidrule(lr){2-9} \cmidrule(lr){10-17} \cmidrule(lr){18-25}
        Map & \(4\) & \(8\) & \(16\) & \(24\) & \(32\) & \(48\) & \(64\) & \(96\) & \(4\) & \(8\) & \(16\) & \(24\) & \(32\) & \(48\) & \(64\) & \(96\) & \(4\) & \(8\) & \(16\) & \(24\) & \(32\) & \(48\) & \(64\) & \(96\) \\
        \midrule
        \(V01\) & \underline{\textbf{0.08}} & \textbf{0.09} & \textbf{0.10} & 0.40 & 0.83 & 1.81 & 4.23 & 10.36 & 2589 & 1451 & 932 & 684 & 607 & \textbf{448} & \textbf{410} & \underline{\textbf{293}} & 52.4 & \underline{\textbf{41.1}} & \textbf{41.9} & \textbf{50.1} & 58.8 & 86.8 & 127.4 & 227.2 \\
        \(V02\) & \textbf{0.09} & \underline{\textbf{0.08}} & \textbf{0.08} & 0.39 & 0.77 & 1.67 & 3.54 & 8.82 & 1996 & 910 & 454 & 318 & 248 & \textbf{174} & \textbf{144} & \underline{\textbf{101}} & 51.1 & \underline{\textbf{38.8}} & \textbf{39.0} & \textbf{47.1} & 56.0 & 84.0 & 124.9 & 224.7 \\
        \(V03\) & \textbf{0.08} & \underline{\textbf{0.08}} & \textbf{0.08} & 0.40 & 0.72 & 1.61 & 3.15 & 8.04 & 1334 & 620 & 261 & 193 & 139 & \textbf{99.84} & \textbf{69.20} & \underline{\textbf{50.38}} & 50.9 & \textbf{38.4} & \underline{\textbf{38.3}} & \textbf{46.4} & 55.2 & 83.3 & 124.2 & 224.0 \\
        \(V04\) & \textbf{0.08} & \underline{\textbf{0.08}} & \textbf{0.08} & 0.40 & 0.70 & 1.52 & 2.91 & 7.69 & 856 & 363 & 161 & 125 & 81.32 & \textbf{55.26} & \textbf{38.26} & \underline{\textbf{30.50}} & 50.7 & \textbf{38.0} & \underline{\textbf{37.9}} & \textbf{45.9} & 54.8 & 82.8 & 123.8 & 223.7 \\
        \(V05\) & \textbf{0.08} & \textbf{0.08} & \underline{\textbf{0.07}} & 0.37 & 0.69 & 1.38 & 2.63 & 6.14 & 172 & 95.56 & 45.92 & 29.29 & 21.07 & \textbf{13.34} & \textbf{8.29} & \underline{\textbf{5.59}} & 50.5 & \textbf{37.7} & \underline{\textbf{37.5}} & \textbf{45.5} & 54.3 & 82.4 & 123.4 & 223.3 \\
        \(V06\) & \underline{\textbf{0.06}} & \textbf{0.07} & \textbf{0.07} & 0.35 & 0.55 & 1.21 & 1.92 & 5.12 & 181 & 86.30 & 39.19 & 29.74 & 20.61 & \textbf{17.04} & \textbf{12.18} & \underline{\textbf{8.97}} & 49.9 & \underline{\textbf{37.8}} & \textbf{38.0} & \textbf{46.1} & 55.0 & 83.1 & 124.0 & 223.9 \\
        \midrule
        \textbf{Ratio} (\(\times\)) & \multicolumn{8}{c|}{Preprocessing} & \multicolumn{8}{c|}{Query speedup} & \multicolumn{8}{c}{Index memory} \\
        \midrule
        \(V01\) & 0.04 & 0.07 & 0.12 & 0.44 & 0.62 & \textbf{0.82} & \textbf{0.89} & \underline{\textbf{0.96}} & \underline{\textbf{1582}} & \textbf{1087} & \textbf{680} & 555 & 502 & 363 & 311 & 207 & 0.24 & 0.33 & 0.49 & 0.61 & 0.70 & \textbf{0.82} & \textbf{0.90} & \underline{\textbf{0.96}} \\
        \(V02\) & 0.06 & 0.09 & 0.12 & 0.47 & 0.65 & \textbf{0.82} & \textbf{0.90} & \underline{\textbf{0.96}} & \underline{\textbf{1121}} & \textbf{669} & \textbf{391} & 270 & 223 & 149 & 112 & 82.9 & 0.24 & 0.32 & 0.48 & 0.59 & 0.69 & \textbf{0.82} & \textbf{0.90} & \underline{\textbf{0.96}} \\
        \(V03\) & 0.06 & 0.11 & 0.14 & 0.51 & 0.67 & \textbf{0.86} & \textbf{0.91} & \underline{\textbf{0.96}} & \underline{\textbf{706}} & \textbf{403} & \textbf{202} & 140 & 118 & 75.5 & 61.7 & 36.9 & 0.24 & 0.32 & 0.47 & 0.59 & 0.69 & \textbf{0.81} & \textbf{0.90} & \underline{\textbf{0.96}} \\
        \(V04\) & 0.08 & 0.13 & 0.16 & 0.55 & 0.70 & \textbf{0.87} & \textbf{0.93} & \underline{\textbf{0.97}} & \underline{\textbf{469}} & \textbf{231} & \textbf{127} & 93.8 & 69.3 & 45.5 & 31.1 & 20.2 & 0.23 & 0.31 & 0.47 & 0.58 & 0.69 & \textbf{0.81} & \textbf{0.90} & \underline{\textbf{0.96}} \\
        \(V05\) & 0.10 & 0.16 & 0.18 & 0.60 & 0.75 & \textbf{0.89} & \textbf{0.94} & \underline{\textbf{0.98}} & \underline{\textbf{127}} & \textbf{69.9} & \textbf{43.6} & 29.6 & 19.9 & 14.6 & 8.16 & 6.44 & 0.23 & 0.31 & 0.47 & 0.58 & 0.69 & \textbf{0.81} & \textbf{0.90} & \underline{\textbf{0.96}} \\
        \(V06\) & 0.07 & 0.14 & 0.18 & 0.56 & 0.72 & \textbf{0.87} & \textbf{0.91} & \underline{\textbf{0.96}} & \underline{\textbf{103}} & \textbf{61.8} & \textbf{29.8} & 22.5 & 17.5 & 13.6 & 10.1 & 7.34 & 0.23 & 0.31 & 0.47 & 0.59 & 0.69 & \textbf{0.81} & \textbf{0.90} & \underline{\textbf{0.96}} \\
        \bottomrule
    \end{tabular}}
    \end{adjustwidth}
\end{table}

To gauge the contribution of the recursive hierarchy, Table~\ref{tab:flat-map} (by map) and Table~\ref{tab:flat-regime} (by regime) report single-level Flat BRICK (\texttt{BrickSearch}) at \(g=4\), which materializes identical base-tile Kleene closures \((A_T^\star, R_T^{VB}, S_T, R_T^{BV})\) but resolves inter-tile reachability via online breadth-first search (BFS) over the flat boundary-port graph rather than precomputed hierarchical projections.
Entries in these tables are averaged over the six regimes or over the eight maps respectively, and the H-BRICK reference values use the same averaging.
At the operating point \((b=24, g=4)\), the recursive hierarchy reduces mean query latency by \(21.5\times\) on \Map{1}, by \(24.3\times\) on \Map{3}, and by \(31.0\times\) to \(381.7\times\) on the remaining maps---from \(152.32\,\mu\mathrm{s}\) to \(1.19\,\mu\mathrm{s}\) on \Map{4} (\(127.6\times\)) and from \(1{,}186.40\,\mu\mathrm{s}\) to \(3.11\,\mu\mathrm{s}\) on \Map{8} (\(381.7\times\))---while requiring \(1.49\times\) to \(4.42\times\) more preprocessing (\(0.330\,\mathrm{s}\) vs.\ \(0.075\,\mathrm{s}\) on \Map{4}) and \(1.07\times\) to \(1.87\times\) more index memory.
The mechanism behind the gap is structural: flat port-graph BFS visits a number of ports that grows with the reachable region of the source, whereas the hierarchical query touches a bounded number of ancestor interfaces regardless of how far the reachable region extends.
The base-tile sweep on \Map{8} (Table~\ref{tab:flat-map}) highlights the scaling barrier of single-level decomposition: fine tiles (\(b \le 8\)) turn most passable cells into boundary ports and inflate flat port-graph BFS traversal to \(5{,}783.81\,\mu\mathrm{s}\), whereas coarse tiles (\(b=96\)) lower search latency to \(436.12\,\mu\mathrm{s}\) at the cost of quadratic intra-tile closure explosion (\(29.16\,\mathrm{s}\) preprocessing, \(793.3\,\mathrm{MiB}\) index).
\texttt{H-BRICK}, by contrast, stays between \(2.90\) and \(3.75\,\mu\mathrm{s}\) for every \(b\), so that its advantage over flat search grows from \(116\times\) at \(b=96\) to \(1{,}657\times\) at \(b=4\); this insensitivity to the size of the flat boundary graph is the behaviour the recursive hierarchy is designed to deliver. Across orientation regimes (Table~\ref{tab:flat-regime}), the query ratio at \(b=24\) falls from \(555\times\) under \(V01\) to \(22.5\times\) under \(V06\), tracking the shrinking reachable regions, while the preprocessing and memory ratios remain within \(0.44\)--\(0.60\) and \(0.58\)--\(0.61\).

\subsection{Batched Many-Source/Many-Target Queries}
\label{subsec:batch-performance}

\begin{figure*}[h]
    \centering
    \setlength{\abovecaptionskip}{0pt}
    \includegraphics[width=\textwidth]{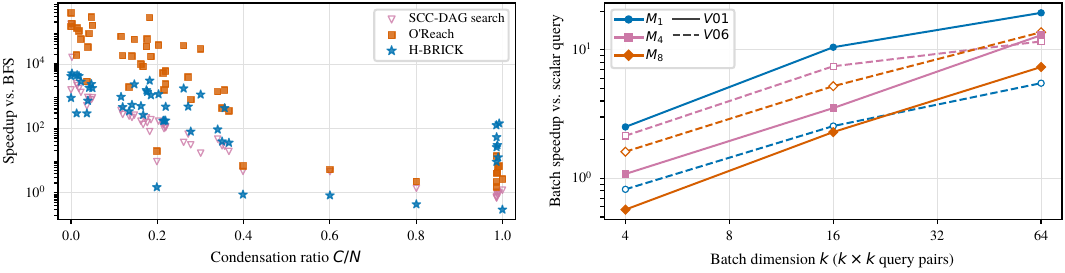}
    \caption{Condensation levels and batch sharing across the benchmark recipes (higher is better).
        (a) Speedup over \texttt{BFS} against the SCC condensation ratio \(C/N\), where \(C\) and \(N\) denote the numbers of strongly connected components and passable vertices; lower ratios indicate stronger graph contraction.
        (b) Batched \texttt{H-BRICK} speedup over scalar query evaluation against the batch dimension \(k\) evaluating \(k \times k\) Cartesian query pairs, with solid curves for \(V01\) and dashed curves for \(V06\).}
    \label{fig:scc_batch}
\end{figure*}

In multi-robot coordination and frontier exploration, reachability queries arrive in batches over common vertex sets \(\mathcal{S} = \{s_1, \dots, s_k\}\) and \(\mathcal{D} = \{t_1, \dots, t_k\}\).
In order to assess the performance of the proposed method in many-to-many reachability check scenarios, Table~\ref{tab:batch-matrix} and Figure~\ref{fig:scc_batch}(b) evaluate the batched query algorithm of Section~\ref{subsec:batch-hbrick} for \(k \in \{4, 16, 64\}\) on all 48 instances.
For each instance and \(k\), \(k\) distinct sources and \(k\) distinct targets are drawn uniformly from the passable vertices.
The batch call is executed twice as warm-up and then eleven times, and its mean time is compared with one scalar pass of \texttt{HBrickSkipLift} over the same \(k \times k\) pairs measured immediately afterwards.
Rather than executing \(k^2\) independent scalar traversals, the batch shares the \(k\) source lifts, the \(k\) target lifts, and the interface reachability frontiers \(\phi_{s_i, U}\) across all pairs meeting at the same ancestor.
The benefit grows with \(k\): at \(k=4\) the batch path is between \(0.49\times\) and \(4.1\times\) the speed of the scalar loop (median \(1.84\times\)), because the shared lifts are amortized over only 16 pairs; at \(k=16\) it is \(2.0\times\)--\(11.4\times\) faster; and at \(k=64\) (\(4{,}096\) pairs) it is \(5.0\times\)--\(22.9\times\) faster (median \(15.5\times\)), reaching \(5.7\,\mathrm{ns}\) per pair on \Map{1} \(V06\) and \(14.5\,\mathrm{ns}\) per pair on \Map{3} \(V01\) (Table~\ref{tab:batch-matrix}).
External graph reachability indices such as \texttt{O'Reach} and \texttt{GRAIL} lack batch work-sharing primitives and require \(k^2\) independent scalar queries.
In multi-agent fleet dispatch, where thousands of agent-to-goal queries are evaluated per cycle, batched \texttt{H-BRICK} resolves the complete \(4{,}096\)-pair Cartesian workload in \(0.39\,\mathrm{ms}\) on \Map{8} under \(V06\) (\(96\,\mathrm{ns}\) amortized per pair) and in \(1.57\,\mathrm{ms}\) under \(V01\) (\(383\,\mathrm{ns}\) per pair), where the giant component makes almost every pair reachable and leaves less work to share.

\begin{table}[h]
    \caption{Batched speedup over the scalar \texttt{HBrickSkipLift} loop for \(k=4\) (16 pairs), \(k=16\) (256 pairs), and \(k=64\) (4{,}096 pairs), reported in each cell as \(k=4\), \(k=16\), \(k=64\), for every benchmark map (rows) and orientation regime (columns). All 144 batch evaluations were verified against \texttt{BFS} with zero mismatches.}
    \label{tab:batch-matrix}
    \centering
    \scriptsize
    \setlength{\tabcolsep}{2.5pt}
    \renewcommand{\arraystretch}{1.12}
    \begin{tabular*}{\linewidth}{@{\extracolsep{\fill}}c r|r|r|r|r|r@{}}
        \toprule
        Map & \(V01\) & \(V02\) & \(V03\) & \(V04\) & \(V05\) & \(V06\) \\
        \midrule
        \Map{1} & \makebox[2.4em][r]{2.5}, \makebox[2.4em][r]{10.5}, \makebox[2.4em][r]{19.4} & \makebox[2.4em][r]{0.6}, \makebox[2.4em][r]{2.7}, \makebox[2.4em][r]{7.1} & \makebox[2.4em][r]{0.6}, \makebox[2.4em][r]{2.4}, \makebox[2.4em][r]{5.5} & \makebox[2.4em][r]{0.6}, \makebox[2.4em][r]{2.0}, \makebox[2.4em][r]{5.0} & \makebox[2.4em][r]{0.5}, \makebox[2.4em][r]{2.2}, \makebox[2.4em][r]{5.1} & \makebox[2.4em][r]{0.8}, \makebox[2.4em][r]{2.5}, \makebox[2.4em][r]{5.5} \\
        \Map{2} & \makebox[2.4em][r]{1.6}, \makebox[2.4em][r]{4.6}, \makebox[2.4em][r]{17.0} & \makebox[2.4em][r]{1.9}, \makebox[2.4em][r]{6.8}, \makebox[2.4em][r]{19.6} & \makebox[2.4em][r]{2.2}, \makebox[2.4em][r]{7.4}, \makebox[2.4em][r]{22.9} & \makebox[2.4em][r]{2.8}, \makebox[2.4em][r]{8.2}, \makebox[2.4em][r]{19.8} & \makebox[2.4em][r]{2.2}, \makebox[2.4em][r]{4.8}, \makebox[2.4em][r]{18.3} & \makebox[2.4em][r]{2.9}, \makebox[2.4em][r]{6.3}, \makebox[2.4em][r]{14.3} \\
        \Map{3} & \makebox[2.4em][r]{1.8}, \makebox[2.4em][r]{7.6}, \makebox[2.4em][r]{17.1} & \makebox[2.4em][r]{1.9}, \makebox[2.4em][r]{11.4}, \makebox[2.4em][r]{14.5} & \makebox[2.4em][r]{1.9}, \makebox[2.4em][r]{9.2}, \makebox[2.4em][r]{17.8} & \makebox[2.4em][r]{1.8}, \makebox[2.4em][r]{8.5}, \makebox[2.4em][r]{15.6} & \makebox[2.4em][r]{1.8}, \makebox[2.4em][r]{6.8}, \makebox[2.4em][r]{15.4} & \makebox[2.4em][r]{1.3}, \makebox[2.4em][r]{3.0}, \makebox[2.4em][r]{8.6} \\
        \Map{4} & \makebox[2.4em][r]{1.1}, \makebox[2.4em][r]{3.5}, \makebox[2.4em][r]{13.0} & \makebox[2.4em][r]{1.4}, \makebox[2.4em][r]{4.6}, \makebox[2.4em][r]{10.9} & \makebox[2.4em][r]{0.8}, \makebox[2.4em][r]{5.2}, \makebox[2.4em][r]{14.4} & \makebox[2.4em][r]{2.2}, \makebox[2.4em][r]{4.2}, \makebox[2.4em][r]{14.9} & \makebox[2.4em][r]{2.4}, \makebox[2.4em][r]{7.5}, \makebox[2.4em][r]{12.8} & \makebox[2.4em][r]{2.1}, \makebox[2.4em][r]{7.5}, \makebox[2.4em][r]{11.6} \\
        \Map{5} & \makebox[2.4em][r]{1.9}, \makebox[2.4em][r]{5.7}, \makebox[2.4em][r]{20.4} & \makebox[2.4em][r]{1.6}, \makebox[2.4em][r]{6.5}, \makebox[2.4em][r]{18.0} & \makebox[2.4em][r]{1.4}, \makebox[2.4em][r]{8.1}, \makebox[2.4em][r]{17.6} & \makebox[2.4em][r]{2.4}, \makebox[2.4em][r]{7.4}, \makebox[2.4em][r]{17.3} & \makebox[2.4em][r]{2.0}, \makebox[2.4em][r]{7.6}, \makebox[2.4em][r]{18.7} & \makebox[2.4em][r]{2.5}, \makebox[2.4em][r]{7.2}, \makebox[2.4em][r]{19.9} \\
        \Map{6} & \makebox[2.4em][r]{1.7}, \makebox[2.4em][r]{6.5}, \makebox[2.4em][r]{20.8} & \makebox[2.4em][r]{1.9}, \makebox[2.4em][r]{5.7}, \makebox[2.4em][r]{17.5} & \makebox[2.4em][r]{1.9}, \makebox[2.4em][r]{6.2}, \makebox[2.4em][r]{18.7} & \makebox[2.4em][r]{4.1}, \makebox[2.4em][r]{9.5}, \makebox[2.4em][r]{20.2} & \makebox[2.4em][r]{1.5}, \makebox[2.4em][r]{6.7}, \makebox[2.4em][r]{16.1} & \makebox[2.4em][r]{2.4}, \makebox[2.4em][r]{7.3}, \makebox[2.4em][r]{18.8} \\
        \Map{7} & \makebox[2.4em][r]{1.3}, \makebox[2.4em][r]{4.5}, \makebox[2.4em][r]{13.2} & \makebox[2.4em][r]{2.5}, \makebox[2.4em][r]{4.5}, \makebox[2.4em][r]{17.2} & \makebox[2.4em][r]{1.4}, \makebox[2.4em][r]{6.2}, \makebox[2.4em][r]{16.1} & \makebox[2.4em][r]{2.1}, \makebox[2.4em][r]{6.4}, \makebox[2.4em][r]{14.1} & \makebox[2.4em][r]{2.6}, \makebox[2.4em][r]{10.2}, \makebox[2.4em][r]{12.8} & \makebox[2.4em][r]{2.2}, \makebox[2.4em][r]{6.0}, \makebox[2.4em][r]{16.3} \\
        \Map{8} & \makebox[2.4em][r]{0.6}, \makebox[2.4em][r]{2.3}, \makebox[2.4em][r]{7.3} & \makebox[2.4em][r]{0.5}, \makebox[2.4em][r]{2.3}, \makebox[2.4em][r]{10.1} & \makebox[2.4em][r]{0.8}, \makebox[2.4em][r]{3.9}, \makebox[2.4em][r]{13.0} & \makebox[2.4em][r]{1.4}, \makebox[2.4em][r]{4.8}, \makebox[2.4em][r]{13.2} & \makebox[2.4em][r]{0.8}, \makebox[2.4em][r]{3.5}, \makebox[2.4em][r]{12.8} & \makebox[2.4em][r]{1.6}, \makebox[2.4em][r]{5.2}, \makebox[2.4em][r]{13.7} \\
        \bottomrule
    \end{tabular*}
\end{table}

\begin{table}[th]
    \caption{Break-even query volume \(Q_{\mathrm{BE}}\) of Eq.~\eqref{eq:break-even} and, in parentheses, the corresponding speedup over online BFS, both for \texttt{HBrickSkipLift} at \((b=24,g=4)\), for each orientation regime (rows) and benchmark map (columns). Dashes mark instances where the indexed mean latency does not improve on online BFS, so no finite break-even volume exists.}
    \label{tab:break-even}
    \centering
    \scriptsize
    \setlength{\tabcolsep}{2.6pt}
    \renewcommand{\arraystretch}{1.08}
    \resizebox{\linewidth}{!}{%
        \begin{tabular}{c rrrrrrrr}
            \toprule
            Regime  & \Map{1}              & \Map{2}                   & \Map{3}                   & \Map{4}                 & \Map{5}                  & \Map{6}                  & \Map{7}                  & \Map{8}                  \\
            \midrule
            \(V01\) & 105 (883\(\times\))  & 392 (1950\(\times\))      & 46 (5062\(\times\))       & 263 (1519\(\times\))    & 389 (262\(\times\))      & 260 (670\(\times\))      & 384 (1421\(\times\))     & 364 (4107\(\times\))     \\
            \(V02\) & --- (0.94\(\times\)) & 744 (698\(\times\))       & 46 (4083\(\times\))       & 209 (1697\(\times\))    & 332 (373\(\times\))      & 310 (495\(\times\))      & 433 (1081\(\times\))     & 430 (3823\(\times\))     \\
            \(V03\) & --- (0.64\(\times\)) & 1{,}291 (358\(\times\))   & 58 (3549\(\times\))       & 183 (1593\(\times\))    & 468 (244\(\times\))      & 321 (319\(\times\))      & 526 (716\(\times\))      & 722 (1607\(\times\))     \\
            \(V04\) & --- (0.58\(\times\)) & 2{,}998 (192\(\times\))   & 81 (2213\(\times\))       & 272 (882\(\times\))     & 683 (147\(\times\))      & 945 (149\(\times\))      & 880 (411\(\times\))      & 979 (959\(\times\))      \\
            \(V05\) & --- (0.27\(\times\)) & 8{,}955 (67.8\(\times\))  & 633 (276\(\times\))       & 578 (370\(\times\))     & 3{,}474 (34.5\(\times\)) & 4{,}793 (29.4\(\times\)) & 4{,}103 (103\(\times\))  & 3{,}346 (427\(\times\))  \\
            \(V06\) & --- (0.38\(\times\)) & 57{,}581 (10.1\(\times\)) & 21{,}674 (11.3\(\times\)) & 1{,}085 (169\(\times\)) & 3{,}951 (25.3\(\times\)) & 4{,}078 (29.7\(\times\)) & 5{,}780 (53.0\(\times\)) & 7{,}213 (98.6\(\times\)) \\
            \bottomrule
        \end{tabular}}
\end{table}

\subsection{Break-Even Query Threshold}
\label{subsec:break-even}

Because \texttt{H-BRICK} incurs an upfront preprocessing cost to construct the hierarchy, this build time is amortized over repeated queries.
The break-even query threshold
\begin{equation}
    \label{eq:break-even}
    Q_{\mathrm{BE}} = \frac{T_{\mathrm{pre}}}{T_{\mathrm{search}} - \bar T_{\mathrm{query}}}
\end{equation}
defines the query volume at which cumulative indexed query time amortizes offline hierarchy construction, where \(T_{\mathrm{pre}}\) denotes offline index build time, \(T_{\mathrm{search}}\) is the online traversal latency of unindexed \texttt{BFS}, and \(\bar T_{\mathrm{query}}\) is the mean query latency of \texttt{H-BRICK}.
Table~\ref{tab:break-even} reports, for \texttt{HBrickSkipLift} at \((b=24,g=4)\), the threshold for every regime and map together with the corresponding speedup over \texttt{BFS}: under \(V01\) break-even is reached after only \(46\) queries on \Map{3} and \(105\) on \Map{1}, and between \(46\) and \(392\) across the eight maps; in sparse-flow regimes the volume grows by orders of magnitude, up to \(57{,}581\) queries on \Map{2} under \(V06\), and on \Map{1} no finite threshold exists from \(V02\) onward because the indexed latency no longer improves on \texttt{BFS}.
In persistent multi-query settings such as multi-agent pathfinding and robotic dispatching, where repeated queries are evaluated over static grid layouts, \texttt{H-BRICK} operates deeply within the amortized regime.
Evaluating the same pairs in batches of \(k=64\) lowers the per-pair latency by a factor of \(5.0\) to \(22.9\) relative to one scalar \texttt{HBrickSkipLift} pass over those pairs (median \(15.5\); Table~\ref{tab:batch-matrix}).
The break-even count itself falls by only about \(0.2\%\) on a typical instance, because the \texttt{BFS} latency and the scalar \texttt{H-BRICK} query in the denominator of \eqref{eq:break-even} already differ by a few orders of magnitude affecting the break-even query count insignificantly.
However, the wall-clock time spent answering that same volume shortens by the batch factor, \(5.0\times\) to \(22.9\times\).

\subsection{External Baseline Comparison on Benchmark Workloads}
\label{subsec:exactness}
\label{subsec:memory-scaling}
\label{subsec:compared-methods}
\label{subsec:external-baselines}
\label{subsec:query-performance}
\label{subsec:stability}

\begin{table}[!t]
    \caption{Median of the eight map-level mean query latencies, in microseconds, for each orientation regime, with the best per-regime value in bold (lower is better); \texttt{H-BRICK} is \texttt{HBrickSkipLift} at \((b=24,g=4)\). \emph{Overall} is the geometric mean over the 48 matched per-instance \texttt{BFS} speedups. \texttt{TwoHop} cells are dashes because it completed 11 of 48 instances and no regime reaches full coverage.}
    \label{tab:external-regime-summary}
    \centering
    \scriptsize
    \setlength{\tabcolsep}{6pt}
    \renewcommand{\arraystretch}{1.1}
    \begin{tabular*}{\textwidth}{@{\extracolsep{\fill}}l r r r r r r r@{}}
        \toprule
        Method & \(V01\) & \(V02\) & \(V03\) & \(V04\) & \(V05\) & \(V06\) & Overall vs \texttt{BFS} \\
        \midrule
        \texttt{BFS} & 1,820 & 1,630 & 1,150 & 509 & 121 & 17.6 & \(\times 1.00\) \\
        \texttt{DFS} & 1,440 & 1,420 & 976 & 423 & 87.4 & 16.5 & \(\times 1.28\) \\
        \texttt{SCC-DAG} & 1.28 & 3.08 & 3.04 & 2.39 & 1.76 & 21.5 & \(\times 73.8\) \\
        \texttt{GRAIL} & 0.296 & 3.13 & 1.61 & 2.28 & 1.37 & 23.9 & \(\times 93.5\) \\
        \texttt{O'Reach} & \textbf{0.012} & \textbf{0.021} & \textbf{0.031} & \textbf{0.0495} & \textbf{0.069} & 4.24 & \(\times 2{,}240\) \\
        \texttt{H-BRICK} & 1.06 & 0.847 & 1.02 & 1.03 & 0.884 & \textbf{0.924} & \(\times 190\) \\
        \texttt{TwoHop} & -- & -- & -- & -- & -- & -- & -- \\
        \bottomrule
    \end{tabular*}
\end{table}

Table~\ref{tab:external-regime-summary} compares \texttt{H-BRICK} at \((b=24,g=4)\) with the external baselines across all six orientation regimes.
Under the random-asymmetric regimes \(V01\)--\(V05\), \texttt{O'Reach} has the lowest mean query latency on every map and is faster than \texttt{H-BRICK} on all \(40\) instances because it computes the strongly connected components, and a query whose endpoints share a component is settled simply by comparing those identifiers.
The largest gap occurs under \(V01\), where the regime-median query latency is \(0.012\,\mu\mathrm{s}\) for \texttt{O'Reach} and \(1.06\,\mu\mathrm{s}\) for \texttt{H-BRICK}; Table~\ref{tab:external-regime-summary} reports the corresponding \texttt{SCC-DAG} and \texttt{GRAIL} measurements.

\begin{table}[t]
    \caption{Query performance under gradient-flow orientation \(V06\) across the benchmark suite. All speedup factors are ratios of mean query latencies relative to online BFS measured in the same campaign and query workload; \texttt{H-BRICK} denotes \texttt{HBrickSkipLift} at \((b=24,g=4)\). The bottom row reports H-BRICK's relative speedup over O'Reach.}
    \label{tab:external_v06}
    \centering
    \scriptsize
    \setlength{\tabcolsep}{4.0pt}
    \renewcommand{\arraystretch}{1.15}
    \begin{tabular*}{\linewidth}{@{\extracolsep{\fill}}l rrrrrrrr@{}}
        \toprule
        Method / Metric & \Map{1} & \Map{2} & \Map{3} & \Map{4} & \Map{5} & \Map{6} & \Map{7} & \Map{8} \\
        \midrule
 \texttt{SCC-DAG} (vs \texttt{BFS}) & \(1.20\times\) & \(0.76\times\) & \(1.04\times\) & \(0.98\times\) & \(1.04\times\) & \(0.88\times\) & \(0.68\times\) & \(0.72\times\) \\
 \texttt{GRAIL} (vs \texttt{BFS}) & \(0.45\times\) & \(0.51\times\) & \(0.46\times\) & \(1.22\times\) & \(1.48\times\) & \(1.30\times\) & \(0.51\times\) & \(1.11\times\) \\
 \texttt{O'Reach} (vs \texttt{BFS}) & \(\mathbf{2.66\times}\) & \(4.02\times\) & \(6.43\times\) & \(6.73\times\) & \(13.8\times\) & \(4.26\times\) & \(2.17\times\) & \(1.53\times\) \\
 \texttt{H-BRICK} (vs \texttt{BFS}) & \(0.38\times\) & \(\mathbf{10.1\times}\) & \(\mathbf{11.3\times}\) & \(\mathbf{169\times}\) & \(\mathbf{25.3\times}\) & \(\mathbf{29.7\times}\) & \(\mathbf{53.0\times}\) & \(\mathbf{98.6\times}\) \\
 Relative vs \texttt{O'Reach} & \(0.14\times\) & \(\mathbf{2.52\times}\) & \(\mathbf{1.75\times}\) & \(\mathbf{25.20\times}\) & \(\mathbf{1.84\times}\) & \(\mathbf{6.98\times}\) & \(\mathbf{24.40\times}\) & \(\mathbf{64.58\times}\) \\
        \bottomrule
    \end{tabular*}
\end{table}

\begin{figure}[H]
    \centering
    \setlength{\abovecaptionskip}{2pt}
    \includegraphics[width=\textwidth]{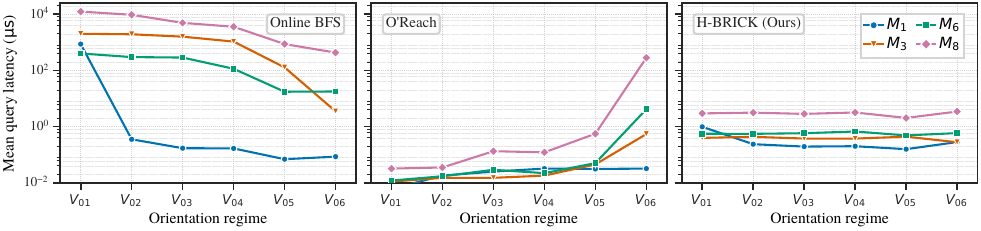}
    \caption{Mean query latency (lower is better) across orientation regimes \(V01\)--\(V06\) for \texttt{BFS}, \texttt{O'Reach}, and \texttt{H-BRICK} (\texttt{HBrickSkipLift} at \(b=32,g=4\)) on four benchmark maps (\Map{1}, \Map{3}, \Map{6}, and \Map{8}). Across these maps, \texttt{H-BRICK} stays in the microsecond range; its max-to-min mean latency is \(1.38\times\) on \Map{6} and \(6.24\times\) on \Map{1}.}
    \label{fig:query_vs_orientation}
\end{figure}

Under gradient flow (\(V06\)) the ordering reverses and \texttt{H-BRICK} is faster than \texttt{O'Reach} on seven of the eight maps, losing only on the corridor maze \Map{1}; Tables~\ref{tab:external-regime-summary} and~\ref{tab:external_v06} also report \texttt{SCC-DAG} and \texttt{GRAIL}, and \(V06\) was evaluated at a single flow angle.
Aggregated over the suite, the geometric-mean speedup over \texttt{BFS} is \(2{,}240\times\) for \texttt{O'Reach} against \(190\times\) for \texttt{H-BRICK}, and \texttt{O'Reach} also builds a smaller index (\(3.7\,\mathrm{MiB}\)) in less time (\(37.8\,\mathrm{ms}\)) than \texttt{H-BRICK} (\(48.5\,\mathrm{MiB}\), \(0.17\,\mathrm{s}\)).
\texttt{H-BRICK} is therefore not the fastest method on every workload; it is a spatially structured exact index whose latency varies less with orientation and that stays in the microsecond range across all six regimes (Figure~\ref{fig:query_vs_orientation}).

\texttt{H-BRICK} admits a batched query that reuses the source and target lifts across many pairs, and at \(k=64\) this is \(5.0\times\)--\(22.9\times\) faster than one scalar pass over the same pairs (Table~\ref{tab:batch-matrix}).
\texttt{O'Reach} has no such sharing, so each pair remains an independent query.
At \(k=64\) under \(V06\), \texttt{H-BRICK} wins on all eight maps and reaches \(34.6\,\mathrm{ns}\) per pair against \(472.7\,\mathrm{ns}\) per pair for \texttt{O'Reach}, a \(13.68\times\) advantage; the batch is thus a complementary use case in which \texttt{H-BRICK} shares work across common endpoints (Table~\ref{tab:batch-oreach-regime}).

\begin{table}[H]
    \caption{Direct batch comparison by orientation regime at \((b=24,g=4)\). Entries are the median mean latency per pair over the eight maps, in nanoseconds (lower is better).}
    \label{tab:batch-oreach-regime}
    \centering
    \scriptsize
    \setlength{\tabcolsep}{4.5pt}
    \renewcommand{\arraystretch}{1.15}
    \begin{tabular*}{\linewidth}{@{\extracolsep{\fill}}l r r r r r r@{}}
        \toprule
        & \(V01\) & \(V02\) & \(V03\) & \(V04\) & \(V05\) & \(V06\) \\
        \midrule
        \texttt{O'Reach} & 3.0 & 3.6 & 6.9 & 11.3 & 10.6 & 472.7 \\
        \texttt{H-BRICK}, \(k=4\) & 709.1 & 715.1 & 444.2 & 404.3 & 315.7 & 265.1 \\
        \texttt{H-BRICK}, \(k=16\) & 273.6 & 166.7 & 140.4 & 89.9 & 61.6 & 68.7 \\
        \texttt{H-BRICK}, \(k=32\) & 144.6 & 95.1 & 63.6 & 61.0 & 45.0 & 49.8 \\
        \texttt{H-BRICK}, \(k=64\) & 81.4 & 54.3 & 41.7 & 35.5 & 26.5 & 34.6 \\
        \midrule
        \(k=64\) vs \texttt{O'Reach} & \(0.04\times\) & \(0.07\times\) & \(0.17\times\) & \(0.32\times\) & \(0.40\times\) & \(13.68\times\) \\
        \bottomrule
    \end{tabular*}
\end{table}

\FloatBarrier
\section{Conclusions}
\label{sec:conclusion}

This paper introduced a hierarchical boundary-compressed Kleene closure for exact directed reachability on grid-embedded graphs.
By projecting regional closures onto perimeter ports and recursively composing interface summaries toward a root, \texttt{H-BRICK} replaces quadratic transitive closure with localized boundary relations algebraically equivalent to original closure projections, ensuring exact reachability queries without approximation error.

The empirical evaluation complements the algebraic exactness result: \texttt{H-BRICK} stores reusable boundary relations in a compositional spatial hierarchy, avoids materializing the global transitive closure, and maintains microsecond-scale mean query latency across the tested orientation regimes.
The observed per-map max-to-min mean-query-latency ratios across the eight benchmark maps range from \(1.38\times\) to \(6.24\times\). At the selected \((b=24, g=4)\) operating point, bit-packed dense transitive closure requires about \(240\times\) the \texttt{H-BRICK} index storage on the largest map.
The hierarchy ablation confirmed the benefit of multi-level composition: at \((b=32, g=4)\), \texttt{H-BRICK} accelerates queries by \(17.0\times\) to \(333.0\times\) over single-level Flat BRICK, with modest preprocessing overhead and \(1.03\times\) to \(1.61\times\) index growth.
Furthermore, batched evaluation demonstrated that sharing boundary lifts and interface projections yields \(5.0\times\)--\(22.9\times\) throughput acceleration over scalar queries at \(k=64\) across the full suite (resolving \(4{,}096\) pairs in under \(2\,\mathrm{ms}\)), with zero reachability mismatches across the entire benchmark suite.
The boundary projection and hierarchical composition depend only on the Kleene star of an idempotent semiring.
The Boolean instance developed here decides reachability; the same scheme yields shortest paths in the tropical min-plus semiring and bottleneck objectives in a min-max semiring.
Realizing those instances, together with natural terrains, 3D voxel grids, dynamic obstacle updates, and distance-bounded reachability, remains future work.

\authorcontributions{Conceptualization and mathematical development, T.Ö. and E.A.Ö.; software and experiments, T.Ö.; writing, T.Ö., with review by E.A.Ö.; supervision, E.A.Ö. All authors have read and agreed to the published version of the manuscript.}

\funding{This work was supported by the Scientific Research Projects Coordination Unit of Ankara Yıldırım Beyazıt University under Multidisciplinary Research Project No.~3225.}

\institutionalreview{Not applicable.}

\informedconsent{Not applicable.}

\dataavailability{The H-BRICK C++20 implementation, benchmark harnesses, scripts, and reproduction recipes are available at \url{https://github.com/ozaslan/hbrick} (grid maps from Moving AI \cite{sturtevant2012benchmarks}).}

\conflictsofinterest{The authors declare no conflicts of interest.}

\begin{adjustwidth}{-\extralength}{0cm}

  \reftitle{References}

  \externalbibliography{yes}
  \bibliography{refs}

  \enlargethispage{2\baselineskip}
  \PublishersNote{}
\end{adjustwidth}
\end{document}